\documentclass{article}
\usepackage{iclr2027_conference,times}
\usepackage{microtype}
\usepackage{graphicx}
\usepackage{needspace}
\usepackage{float,wrapfig,placeins}
\usepackage{amsmath,amssymb,mathtools,amsthm}
\usepackage{booktabs,makecell,array,tabularx,multirow}
\usepackage{xcolor,colortbl,siunitx,pifont}
\usepackage{fontawesome5}
\usepackage[skins,breakable]{tcolorbox}
\usepackage{hyperref}
\usepackage{url}
\usepackage[capitalize,noabbrev]{cleveref}

\definecolor{colorhead}{HTML}{F0F7EB}
\definecolor{colorheadred}{HTML}{F9EEEC}
\definecolor{colorrow}{HTML}{EDEDEB}
\newsavebox{\filltabbox}
\newlength{\filltabsep}
\newcommand{\filltabular}[3]{%
  \begingroup
  \setlength{\tabcolsep}{0pt}%
  \sbox{\filltabbox}{\begin{tabular}{#2}#3\end{tabular}}%
  \setlength{\filltabsep}{\dimexpr(\linewidth-\wd\filltabbox)/(2*#1)\relax}%
  \setlength{\tabcolsep}{\filltabsep}%
  \begin{tabular}{#2}#3\end{tabular}%
  \endgroup}
\definecolor{visblue}{RGB}{50,105,210}
\definecolor{textorange}{RGB}{220,120,30}
\definecolor{gainblue}{HTML}{245E87}

\definecolor{lossred}{HTML}{B83A2E}

\definecolor{findinggreen}{HTML}{3F6846}
\definecolor{findingback}{HTML}{EAF3E6}
\definecolor{revblue}{HTML}{1A56DB}
\newif\ifrevmarkup
\revmarkupfalse
\ifrevmarkup
  \newcommand{\rev}[1]{\textcolor{revblue}{#1}}
\else
  \newcommand{\rev}[1]{#1}
\fi

\newcommand{\score}[2]{#1\if\relax\detokenize{#2}\relax\else\textcolor{blue}{{\scriptsize\,(#2)}}\fi}
\newcommand{\scoregray}[2]{#1\if\relax\detokenize{#2}\relax\else\textcolor{gray}{{\scriptsize\,(#2)}}\fi}
\newcommand{\scorered}[2]{#1\if\relax\detokenize{#2}\relax\else\textcolor{red}{{\scriptsize\,(#2)}}\fi}

\newtcolorbox{hypothesisbox}{%
  enhanced,
  colback=findingback,
  frame hidden,
  borderline west={1.8pt}{0pt}{findinggreen},
  arc=0pt,
  left=6.5pt,
  right=5pt,
  top=3pt,
  bottom=3pt,
  boxsep=0pt,
  before skip=5.5pt,
  after skip=5.5pt
}
\newcommand{\hypobox}[3]{%
  \begin{hypothesisbox}
    \textcolor{findinggreen}{\scriptsize\bfseries\MakeUppercase{#1}}%
    \hspace{3pt}\small\textbf{#2}\enspace\emph{#3}
  \end{hypothesisbox}%
}

\DeclareRobustCommand{\Vis}{%
  \ensuremath{{\mathcal{\textcolor{blue}{V}}}%
  }}

\DeclareRobustCommand{\Txt}{%
  \ensuremath{{\mathcal{\textcolor{orange}{T}}}%
  }}

\DeclareRobustCommand{\Rblk}{%
  \ensuremath{{\mathcal{R}}%
  }}

\DeclareRobustCommand{\BaseFmt}{\ensuremath{\langle \Vis, \Txt \rangle}}
\DeclareRobustCommand{\SLPFmt}{\ensuremath{\langle \Vis, \Txt, \Vis, \Txt \rangle}}

\DeclareRobustCommand{\TVFmt}{\ensuremath{\langle \Txt, \Vis \rangle}}
\DeclareRobustCommand{\TVTFmt}{\ensuremath{\langle \Txt, \Vis, \Txt \rangle}}
\DeclareRobustCommand{\VVTFmt}{\ensuremath{\langle \Vis, \Vis, \Txt \rangle}}
\DeclareRobustCommand{\VTVFmt}{\ensuremath{\langle \Vis, \Txt, \Vis \rangle}}
\DeclareRobustCommand{\VTTFmt}{\ensuremath{\langle \Vis, \Txt, \Txt \rangle}}
\DeclareRobustCommand{\VVTTFmt}{\ensuremath{\langle \Vis, \Vis, \Txt, \Txt \rangle}}

\title{Replay Before You Reply: Second-Look\\Prompting for Unified Multimodal Encoding}

\iclrfinalcopy

\definecolor{linkblue}{rgb}{0.21,0.49,0.74}
\newcommand{\projectpage}{\multicolumn{1}{@{}l@{}}{\makebox[\textwidth][c]{%
  \fontsize{12}{16}\selectfont\hypersetup{pdfborder={0 0 0}}%
  \href{https://second-look-vlm.com/}{\faGlobe\ \textcolor{linkblue}{Second-Look Prompting Project Page}}}}}
\author{Yicheng Zhang$^{1}$\thanks{Equal contribution. Contact: \texttt{zhangyic23@mails.tsinghua.edu.cn}.}\hphantom{$^{*}$}\quad
Xun Zhang$^{1,2*}$\quad Jingyuan Ma$^{1*}$\quad
Tianyu Yu$^{1}$\thanks{Corresponding authors: \texttt{yiranytianyu@gmail.com}, \texttt{yaoyuanthu@gmail.com}.}\hphantom{$^{\dagger}$}\quad
Yuan Yao$^{1\dagger}$\\[0.2cm]
$^{1}$Tsinghua University, China\quad $^{2}$Shanghai Jiao Tong University, China\\[0.1cm]
\projectpage}

\begin{document}
\maketitle
\vspace{-0.2in}
\lhead{Preprint}
\suppressfloats[t]
\begin{abstract}
Solving multimodal tasks requires Vision-Language Models (VLMs) to form a unified understanding of the image and the question text. However, under causal attention in current VLMs, a token can attend only to what precedes it. Whichever modality comes first is therefore encoded under partial observation of the input, uninformed by the rest of the multimodal context. We introduce Second-Look Prompting (SLP), a training-free method that repeats the image--question pair in the input. In the repeated copy, every visual and textual token can attend to the complete input, yielding a unified multimodal encoding of the pair. Moreover, the added visual tokens contribute computation beyond re-exposing the image---even a blank image in their place can improve reasoning. Across 12 benchmarks, SLP improves the overall average score of every evaluated model, spanning nine open-source models (3B--72B) and three proprietary frontier models, with gains reaching 5.6 points for Gemini 3 Flash on MATH-Vision. Our analysis reveals two complementary mechanisms: the question guides where the second image attends, and the added visual tokens act as a prefill scratchpad whose hidden states retain answer-relevant information from the image and question. These findings highlight input presentation as a simple yet effective way to improve VLM performance.

\end{abstract}

\newlength{\savedtextfloatsep}
\setlength{\savedtextfloatsep}{\textfloatsep}
\setlength{\textfloatsep}{10pt plus 2pt minus 2pt}
\section{Introduction}\label{sec:intro}

Vision-Language Models (VLMs)~\citep{qwen2.5-VL,team2025gemma,li2024llavaonevision} answer questions about images by encoding the image and the question as a single causal token sequence. Prior work improves their reasoning through generation-time strategies~\citep{wei2022chain,wang2022self,zhang2023multimodal}; we instead ask how the input presentation shapes representations during prefill. Under causal attention, a single-pass presentation in either order allows only one-way cross-modal interaction: the earlier modality cannot attend to the later one. In the standard image-before-question prompt, denoted \BaseFmt{}, visual tokens therefore cannot attend to the question. Here, \Vis{} denotes the image block and \Txt{} the complete question text, including instructions and options. We refer to this constraint as the \emph{cross-modal partial-observation problem}.

\begin{figure}[!t]
\centering
\includegraphics[width=\textwidth]{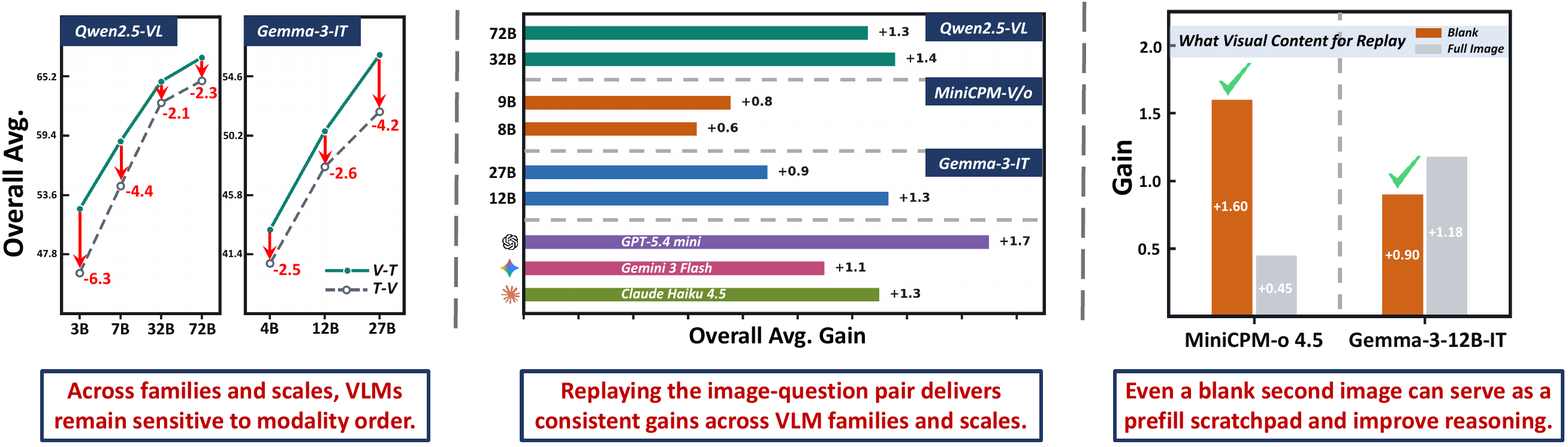}
\caption{\textbf{Motivation, broad gains, and scratchpad evidence for Second-Look Prompting.} Left: across multiple scales in both the Qwen2.5-VL and Gemma-3-IT families, reversing the standard \BaseFmt{} order to \TVFmt{} lowers the overall average by 2.1--6.3 points, showing that VLMs remain sensitive to input order. Middle: relative to \BaseFmt{}, SLP consistently improves the overall average of representative open-source and proprietary VLMs. Right: averaged over seven reasoning benchmarks, the bars compare inserting a blank image, defined as \(\langle\Vis,\Txt,\square,\Txt\rangle\), or replaying the same image, \SLPFmt{}, both relative to text-only replay, \VTTFmt{}. The positive blank-image gains suggest that added visual positions can support reasoning without adding task-relevant pixels.}
\label{fig:teaser}
\end{figure}

Reversing the prompt to \TVFmt{} merely relocates the partial observation to the question, now encoded without access to the image. Across the tested families and scales, VLM performance remains sensitive to this ordering: \TVFmt{} yields a lower overall average than \BaseFmt{} for every tested model (Figure~\ref{fig:teaser}, left; Appendix~\ref{app:order_reversal}), possibly because the reversed order differs from the training format. Relying on reasoning is also insufficient. In text-only models, rationales re-read the prompt and repetition gains diminish~\citep{leviathan2025promptrepetitionimprovesnonreasoning}; in VLMs, generated reasoning tokens can likewise attend to both modalities, but they cannot restore the missing attention direction (Section~\ref{sec:cot_interaction}).

We introduce \emph{Second-Look Prompting} (SLP), a training-free presentation establishing unified multimodal encoding within a causal backbone: append the same image and question to the standard prompt, forming \SLPFmt{}. During prefill, the second image attends to both the initial image and the question, yielding a question-guided second look; the repeated question then conditions on this updated visual state alongside the full preceding context. SLP requires no parameter updates, added task information, or model-internal access.

Across 12 reasoning and perception benchmarks, SLP improves the overall average of all 12 models: nine open-source models from 3B to 72B and three proprietary frontier models. Overall gains reach up to 1.7 points (Figure~\ref{fig:teaser}, middle; Table~\ref{tab:slp-format-total}) and are larger on reasoning tasks on average. Measured end-to-end latency on the nine open-source models changes by at most $\pm0.16$ seconds on family average (Section~\ref{sec:efficiency}). A continued-training study also shows that the model can learn to better use the second look: with matched training data and steps, LLaVA-OneVision-7B-SI gains 1.97 points under SLP versus 0.55 under the standard prompt (Section~\ref{subsec:training-study}).

Three controlled studies examine how the second look supports answering. \textbf{First, the question actively guides the second look.} Once it can attend to the question, the second image directs cross-image attention toward question-relevant regions (Section~\ref{sec:i2_position}). \textbf{Second, added visual tokens can act as a prefill scratchpad.} Their hidden states retain information from the image and question that the model can later use to answer, giving the second look a computational role beyond revisiting the image. For models that effectively write information into a blank second image, placing one between the repeated questions improves reasoning over text-only replay and can even outperform full-image replay (Figure~\ref{fig:teaser}, right; Section~\ref{sec:summarizer}). \textbf{Third, reasoning does not resolve cross-modal partial observation.} Generated text cannot reorganize visual states formed during prefill; empirically, SLP gains remain similar under direct-answer and default prompts, while text-only repetition gains shrink (Section~\ref{sec:cot_interaction}).

Together, these results show that input presentation alone can shape prefill representations and improve VLM performance. This complements work on generation-time reasoning~\citep{zhang2023multimodal,zhou2024image,cao2024dualfocus} and input order~\citep{tan2024order,wardle2024imagefirst,wei2026multimodalorder}. We will release our code.

\providecommand{\SLPName}{Second-Look Prompting}
\providecommand{\SLP}{SLP}

\section{From Causal Attention to Second-Look Prompting}
\label{sec:prelim}

\begin{figure}[htbp]
\centering
\resizebox{\linewidth}{!}{\includegraphics{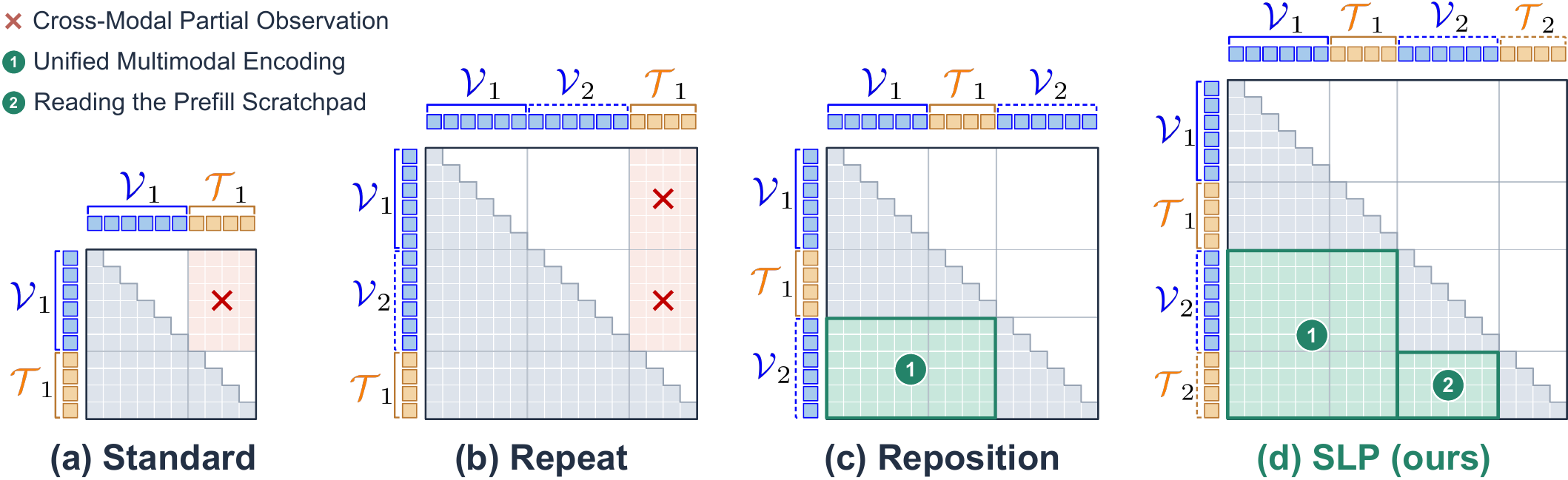}}
\caption{\textbf{Causal attention across input presentations.} Rows are querying tokens and columns are keys; gray and green cells are accessible, blank and red cells are masked. Red crosses mark visual blocks that cannot attend to the question; green cells mark the two steps of SLP described in the text (step~1: question-guided visual encoding; step~2: reading the prefill scratchpad via \(\Txt_2\)). Within-block attention is schematic and may differ by model.}
\label{fig:prelim}
\end{figure}

Figure~\ref{fig:prelim} shows how input presentation shapes cross-modal attention under a causal mask. In the standard prompt \BaseFmt{}, visual tokens precede the question, so their hidden states are computed under partial observation of the input and are never revised. Later generated tokens can read both modalities but only add text states; they cannot produce a question-conditioned visual representation. Creating one requires visual tokens that follow the question, which only the input can provide.

\textbf{What incomplete variants leave out.}
Repetition and repositioning change this picture only partially.
First, intra-modal \emph{repetition} (\VTTFmt{} and \VVTFmt{}; Figure~\ref{fig:prelim}b) expands context within a single modality: repeating text extends question processing~\citep{xu2024re,springer2025repetition,leviathan2025promptrepetitionimprovesnonreasoning}, while repeating the image allows \(\Vis_2\) to attend to \(\Vis_1\). However, because all visual tokens still precede \(\Txt\), they cannot attend to the question, so cross-modal interaction remains one-way.
Second, \emph{repositioning} (\VTVFmt{}; Figure~\ref{fig:prelim}c) places an image copy after the question, allowing \(\Vis_2\) to attend to both \(\Vis_1\) and \(\Txt_1\) during prefill (step 1). However, \(\Txt_1\) precedes \(\Vis_2\) and cannot attend to this question-conditioned visual representation. 

\Needspace{6\baselineskip}
\textbf{Unified multimodal encoding via SLP.}
Second-Look Prompting repeats both modalities:
\[
\BaseFmt{}\quad\longrightarrow\quad
\SLPFmt{} \coloneqq \langle\Vis_1,\Txt_1,\Vis_2,\Txt_2\rangle.
\]
By resolving the partial-observation constraint, \(\Vis_2\) can selectively integrate visual and textual context during prefill. We call \(\Vis_2\) a multimodal \emph{prefill scratchpad}. Appending \(\Txt_2\) then provides text token positions that read this scratchpad together with the full prompt history (step~2) without disturbing the familiar \BaseFmt{} prefix. In the repeated copy, each modality is thus encoded with the other in context. We examine how this scratchpad retrieves information under question guidance in Section~\ref{sec:i2_position} (Finding~1) and evaluate what it retains and computes in Section~\ref{sec:summarizer} (Finding~2).

\section{Experiments}
\label{sec:experiments}

\providecommand{\SLPName}{Second-Look Prompting}
\providecommand{\SLP}{SLP}

\subsection{Experimental Setup}
\label{sec:setup}

\textbf{Models.}
We evaluate \SLP{} on a diverse set of VLM families, covering Qwen2.5-VL~\citep{qwen2.5-VL} at four scales (3B, 7B, 32B, and 72B), Gemma-3-IT~\citep{team2025gemma} at three scales (4B, 12B, and 27B), and MiniCPM-V 4.5 (8B) and MiniCPM-o 4.5 (9B)~\citep{cui2026minicpmo45realtimefullduplex,yu2026minicpmv45cookingefficient,yao2024minicpm}, jointly abbreviated as MiniCPM-V/o 4.5. All open-source models are the official instruction-tuned checkpoints. We also include three proprietary models: GPT-5.4 mini, Gemini 3 Flash, and Claude Haiku 4.5.

\textbf{Benchmarks.}
We group benchmarks into reasoning-oriented and perception-oriented suites. The reasoning suite includes DynaMath~\citep{zou2025dynamath}, MATH-Vision~\citep{wang2024measuring}, LogicVista~\citep{xiao2024logicvista}, VisualPuzzles~\citep{song2025visualpuzzles}, MMMU~\citep{yue2024mmmu}, MathVista~\citep{lu2024mathvista}, and We-Math~\citep{qiao2025we}, while the perception suite includes AI2D~\citep{kembhavi2016diagram}, OCRBench~\citep{liu2024ocrbench}, SEED-Bench-2-Plus~\citep{li2024seed}, MMBench~\citep{liu2024mmbench}, and MMStar~\citep{chen2024we}. MathVista is evaluated on its \texttt{testmini} split and the other benchmarks on their standard evaluation splits; due to inference cost, each proprietary model is evaluated on up to 1,000 samples per benchmark.

\textbf{Protocol.}
Unless otherwise stated, experiments use the default benchmark prompts that permit intermediate reasoning. Model-specific thinking settings and decoding parameters are given in Appendix~\ref{app:prompts_decoding}. Within each comparison, these settings and the prompt template are fixed\rev{, and all reported gains compare presentations on identical examples within each model rather than across model families}. We report benchmark scores on a 0--100 scale and take unweighted means across benchmarks: seven for reasoning, five for perception, and all 12 for the overall average.

\begin{table*}[t]
\centering
\caption{Main results. Shaded rows add \SLP{} to the model above; deltas on the averages are relative to that baseline and computed before rounding; bold marks only the higher suite or overall average within each pair.}
\label{tab:slp-format-total}

\newcommand{\ModelGroupWidth}{2.2cm}
\newcommand{\MetricWidth}{0.75cm}
\newcommand{\AvgWidth}{1.4cm}
\setlength{\tabcolsep}{2pt}
\renewcommand{\arraystretch}{1.12}

\resizebox{\linewidth}{!}{%
\begin{tabular}{
>{\raggedright\arraybackslash}m{\ModelGroupWidth}|
*{7}{>{\centering\arraybackslash}m{\MetricWidth}}
>{\centering\arraybackslash}m{\AvgWidth}|
*{5}{>{\centering\arraybackslash}m{\MetricWidth}}
>{\centering\arraybackslash}m{\AvgWidth}|
>{\centering\arraybackslash}m{\AvgWidth}
}
\toprule
\multirow{2}{\ModelGroupWidth}[-1ex]{\raggedright\arraybackslash\textbf{Model}}
& \multicolumn{8}{c|}{\textbf{Reasoning}}
& \multicolumn{6}{c|}{\textbf{Perception}}
& \multirow{2}{\AvgWidth}[-1ex]{\centering\arraybackslash{\textbf{\textit{Avg.}}}} \\
\cmidrule(lr){2-9} \cmidrule(lr){10-15}
\multicolumn{1}{c|}{}
& {\scriptsize \shortstack{Dyna\\Math}}
& {\scriptsize \shortstack{MATH-\\Vision}}
& {\scriptsize \shortstack{Logic\\Vista}}
& {\scriptsize \shortstack{Visual\\Puzzles}}
& {\scriptsize MMMU}
& {\scriptsize \shortstack{Math\\Vista}}
& {\scriptsize \shortstack{We-\\Math}}
& {\scriptsize \textbf{Avg.}}
& {\scriptsize \shortstack{AI2D}}
& {\scriptsize \shortstack{OCR\\Bench}}
& {\scriptsize \shortstack{SEED-\\Bench-\\2-Plus}}
& {\scriptsize \shortstack{MM\\Bench}}
& {\scriptsize MMStar}
& {\scriptsize \textbf{Avg.}} \\
\midrule
\rowcolor{colorrow} \multicolumn{16}{c}{\textbf{Proprietary}} \\
\midrule
{\scriptsize GPT-5.4 mini} & 80.1 & 48.7 & 63.5 & 46.2 & 70.4 & 68.9 & 88.2 & 66.6 & 84.3 & 79.5 & 72.2 & 84.0 & 65.2 & 77.0 & 70.9\\
\rowcolor{colorhead}{\scriptsize \hspace{0.8em}\textbf{+SLP}} & 80.3 & 50.2 & 63.5 & 46.0 & 71.6 & 74.3 & 89.6 & \score{\textbf{67.9}}{+1.4} & 86.8 & 80.1 & 72.6 & 87.6 & 68.8 & \score{\textbf{79.2}}{+2.1} & \score{\textbf{72.6}}{+1.7}\\
\midrule
{\scriptsize Gemini 3 Flash} & 85.3 & 76.8 & 81.4 & 71.0 & 83.9 & 87.5 & 95.2 & 83.0 & 88.5 & 90.3 & 76.6 & 93.2 & 79.9 & 85.7 & 84.1\\
\rowcolor{colorhead}{\scriptsize \hspace{0.8em}\textbf{+SLP}} & 84.5 & 82.4 & 83.7 & 73.1 & 84.4 & 87.4 & 96.3 & \score{\textbf{84.5}}{+1.5} & 89.3 & 90.1 & 75.7 & 93.6 & 82.1 & \score{\textbf{86.2}}{+0.5} & \score{\textbf{85.2}}{+1.1}\\
\midrule
{\scriptsize Claude Haiku 4.5} & 71.3 & 53.7 & 57.7 & 43.6 & 68.2 & 73.4 & 86.4 & 64.9 & 79.9 & 71.6 & 69.1 & 81.2 & 63.2 & 73.0 & 68.3\\
\rowcolor{colorhead}{\scriptsize \hspace{0.8em}\textbf{+SLP}} & 72.4 & 54.4 & 58.4 & 44.6 & 69.7 & 75.1 & 87.5 & \score{\textbf{66.0}}{+1.1} & 80.6 & 72.4 & 69.2 & 84.0 & 66.0 & \score{\textbf{74.4}}{+1.4} & \score{\textbf{69.5}}{+1.3}\\

\midrule
\rowcolor{colorrow} \multicolumn{16}{c}{\textbf{Open-Source}} \\
\midrule
{\scriptsize Qwen2.5-VL-72B} & 66.7 & 38.6 & 57.5 & 41.6 & 67.7 & 73.5 & 48.8 & 56.3 & 88.6 & 88.6 & 73.7 & 88.2 & 71.0 & 82.0 & 67.0\\
\rowcolor{colorhead}{\scriptsize \hspace{0.8em}\textbf{+SLP}} & 69.0 & 40.5 & 59.7 & 45.0 & 66.7 & 75.4 & 50.5 & \score{\textbf{58.1}}{+1.8} & 88.8 & 89.1 & 73.7 & 89.7 & 71.5 & \score{\textbf{82.6}}{+0.6} & \score{\textbf{68.3}}{+1.3}\\
{\scriptsize Qwen2.5-VL-32B} & 67.0 & 37.3 & 54.4 & 38.2 & 65.8 & 72.7 & 46.9 & 54.6 & 83.6 & 85.5 & 72.6 & 85.5 & 66.8 & 78.8 & 64.7\\
\rowcolor{colorhead}{\scriptsize \hspace{0.8em}\textbf{+SLP}} & 67.5 & 38.8 & 57.7 & 40.8 & 67.7 & 74.9 & 49.4 & \score{\textbf{56.7}}{+2.1} & 84.4 & 85.8 & 71.8 & 86.1 & 67.7 & \score{\textbf{79.2}}{+0.4} & \score{\textbf{66.1}}{+1.4}\\
{\scriptsize Qwen2.5-VL-7B} & 54.2 & 25.4 & 46.5 & 32.9 & 53.8 & 66.6 & 35.3 & 45.0 & 85.1 & 88.3 & 70.9 & 83.2 & 63.9 & \textbf{78.3} & 58.8\\
\rowcolor{colorhead}{\scriptsize \hspace{0.8em}\textbf{+SLP}} & 54.6 & 26.4 & 46.5 & 32.8 & 55.0 & 67.0 & 37.5 & \score{\textbf{45.7}}{+0.7} & 84.6 & 88.3 & 70.7 & 82.4 & 61.5 & \scoregray{77.5}{-0.8} & \score{\textbf{58.9}}{+0.1}\\
{\scriptsize Qwen2.5-VL-3B} & 43.6 & 21.7 & 36.0 & 28.5 & 48.4 & 62.1 & 21.6 & 37.4 & 81.5 & 82.3 & 68.5 & 77.1 & 55.7 & 73.0 & 52.3\\
\rowcolor{colorhead} {\scriptsize \hspace{0.8em}\textbf{+SLP}} & 43.5 & 21.3 & 40.0 & 28.8 & 50.9 & 61.2 & 24.3 & \score{\textbf{38.6}}{+1.1} & 81.8 & 82.9 & 68.9 & 77.1 & 55.6 & \score{\textbf{73.2}}{+0.2} & \score{\textbf{53.0}}{+0.8}\\
\midrule
{\scriptsize MiniCPM-o 4.5} & 61.6 & 46.0 & 51.2 & 16.4 & 69.2 & 72.1 & 57.5 & 53.4 & 80.8 & 81.8 & 65.4 & 84.9 & 66.6 & 75.9 & 62.8\\
\rowcolor{colorhead}{\scriptsize \hspace{0.8em}\textbf{+SLP}} & 62.5 & 46.1 & 54.1 & 15.3 & 70.8 & 74.2 & 57.0 & \score{\textbf{54.3}}{+0.9} & 81.4 & 82.3 & 65.7 & 85.5 & 67.7 & \score{\textbf{76.5}}{+0.6} & \score{\textbf{63.6}}{+0.8}\\
{\scriptsize MiniCPM-V 4.5} & 63.1 & 42.2 & 53.9 & 20.5 & 66.5 & 73.4 & 50.1 & 52.8 & 80.5 & 80.8 & 67.4 & 86.2 & 70.5 & 77.1 & 62.9\\
\rowcolor{colorhead}{\scriptsize \hspace{0.8em}\textbf{+SLP}} & 63.7 & 40.6 & 55.0 & 20.6 & 67.2 & 75.8 & 51.1 & \score{\textbf{53.4}}{+0.6} & 82.5 & 80.9 & 67.7 & 86.6 & 71.0 & \score{\textbf{77.7}}{+0.7} & \score{\textbf{63.6}}{+0.6}\\
\midrule
{\scriptsize Gemma-3-27B-IT} & 50.6 & 36.7 & 47.2 & 15.2 & 57.6 & 60.1 & 42.0 & 44.2 & 83.3 & 75.3 & 70.1 & 79.5 & 56.5 & 72.9 & 56.2\\
\rowcolor{colorhead}{\scriptsize \hspace{0.8em}\textbf{+SLP}} & 48.1 & 37.2 & 45.9 & 16.4 & 59.0 & 62.1 & 46.3 & \score{\textbf{45.0}}{+0.8} & 84.6 & 74.6 & 70.2 & 81.2 & 59.2 & \score{\textbf{73.9}}{+1.0} & \score{\textbf{57.1}}{+0.9}\\
{\scriptsize Gemma-3-12B-IT} & 30.8 & 31.2 & 36.9 & 16.4 & 51.6 & 54.1 & 35.3 & 36.6 & 80.3 & 72.9 & 67.2 & 75.2 & 54.7 & 70.1 & 50.5\\
\rowcolor{colorhead}{\scriptsize \hspace{0.8em}\textbf{+SLP}} & 37.4 & 31.4 & 37.1 & 16.2 & 53.0 & 54.9 & 35.9 & \score{\textbf{38.0}}{+1.4} & 82.2 & 71.1 & 67.4 & 79.2 & 56.8 & \score{\textbf{71.3}}{+1.3} & \score{\textbf{51.9}}{+1.3}\\
{\scriptsize Gemma-3-4B-IT} & 25.9 & 22.2 & 30.7 & 13.8 & 42.0 & 43.8 & 25.7 & 29.1 & 73.5 & 67.9 & 61.5 & 66.1 & 45.9 & 63.0 & 43.2\\
\rowcolor{colorhead}{\scriptsize \hspace{0.8em}\textbf{+SLP}} & 22.6 & 23.9 & 27.3 & 16.0 & 41.9 & 43.0 & 29.7 & \score{\textbf{29.2}}{+0.1} & 74.7 & 65.6 & 61.5 & 69.1 & 46.3 & \score{\textbf{63.5}}{+0.5} & \score{\textbf{43.5}}{+0.2}\\
\bottomrule
\end{tabular}%
}
\end{table*}

\subsection{Main Results}

\textbf{SLP improves open-source VLMs across families and scales.}
Using \SLPFmt{} instead of \BaseFmt{} improves the overall average of all nine open-source models, spanning 3B to 72B, by $+0.8$ points on average. The gains are generally larger on the reasoning suite: \SLP{} raises the reasoning averages of Qwen2.5-VL-32B and -72B by 2.1 and 1.8 points, against 0.4 and 0.6 on perception. The reasoning-suite average improves on all nine models and the perception average on eight; only Qwen2.5-VL-7B's perception average declines ($-0.8$). \rev{Additional evaluations on recent open-source releases (Qwen3.5-35B-A3B, Gemma-4-E4B-it, and Qwen3.8-27B across four reasoning modes) yield even larger average gains of $+1.6$ to $+4.0$ points (Appendix~\ref{app:new_models}).}

\textbf{SLP also improves proprietary VLMs.}
\SLP{} improves the overall averages of GPT-5.4 mini, Gemini 3 Flash, and Claude Haiku 4.5 by $+1.7$, $+1.1$, and $+1.3$ points, and all three improve on both suites. A paired bootstrap over the 12 benchmarks gives 95\% confidence intervals excluding zero for the overall gains of nine of the 12 models, including all three proprietary models (Appendix~\ref{app:significance}).

\subsection{Dissecting the Second Look}

We ask three questions about the second look: where the second image attends during prefill (Section~\ref{sec:i2_position}), what role its hidden states play in answering (Section~\ref{sec:summarizer}), and whether generated reasoning can substitute for prefill replay (Section~\ref{sec:cot_interaction}).

\subsubsection{Where Does the Second Image Look?}
\label{sec:i2_position}

\hypobox{Finding 1}{The question can guide where second-image tokens attend.}{With the full input in context, second-image attention responds to what is asked.}

Figure~\ref{fig:teaser} (left) shows that VLM performance is sensitive to the order of the image and question. This motivates comparing where to place a single image replay: before the question (\VVTFmt{}) or after it (\VTVFmt{}). Both formats contain identical inputs, differing only in the position of \(\Vis_2\) (Table~\ref{tab:replay_variants}).

\begin{figure}[t]
\centering
\includegraphics[width=\linewidth]{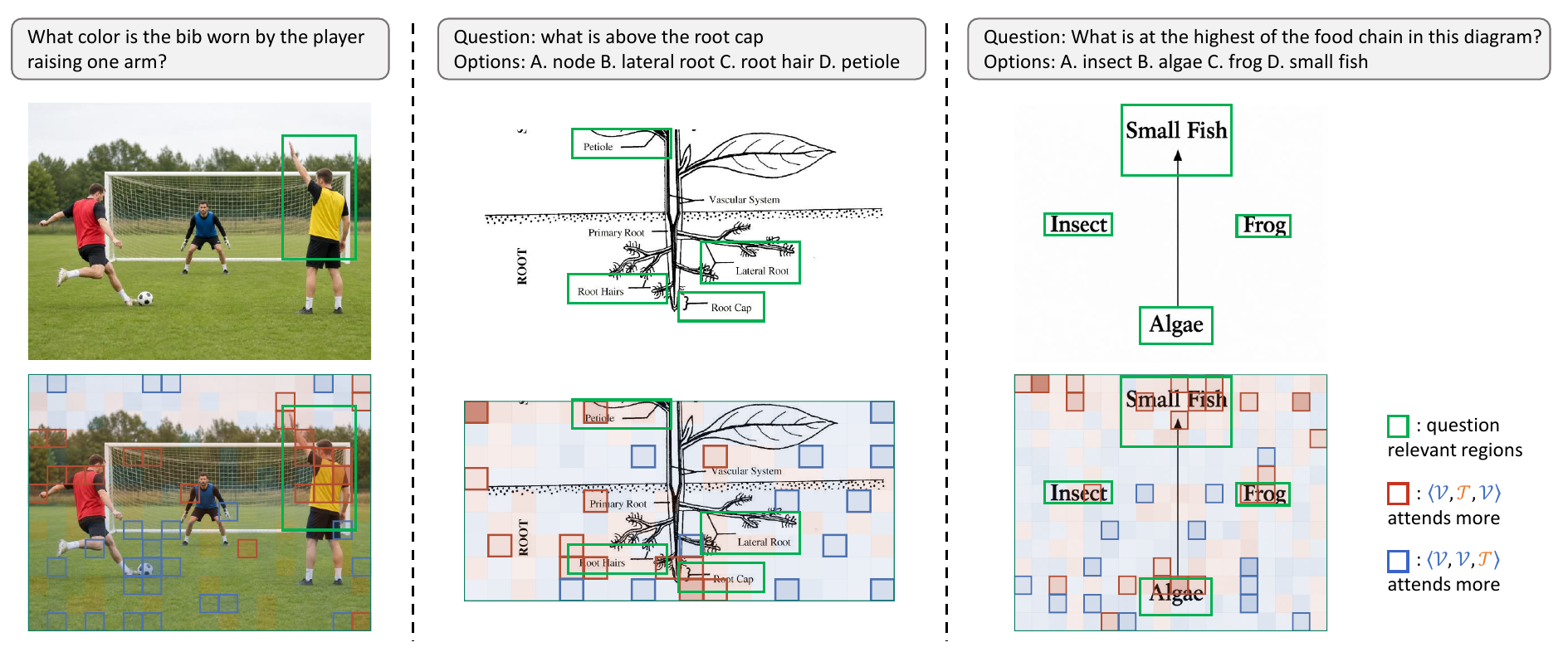}
\caption{Three examples of attention from the second image to the first. Attention is normalized within the first image before averaging over second-image patch queries; we visualize the difference (\VTVFmt{} $-$ \VVTFmt{}). Green boxes mark question-relevant regions. Red and blue outlines mark the smallest set of patches accounting for 50\% of the total absolute difference (red: \VTVFmt{} higher; blue: \VVTFmt{} higher).}
\label{fig:dynamath_sample}
\end{figure}

\Needspace{8\baselineskip}
\noindent\begin{minipage}[t]{0.47\linewidth}
\vspace{0pt}
\begingroup\hyphenpenalty=5000\exhyphenpenalty=5000
Post-question placement achieves higher gains in two of the three model families and on average ($+0.5$ vs.\ $+0.4$), despite a small reversal on MiniCPM-V/o 4.5. Repeating only the question (\VTTFmt{}) yields a smaller average gain of $+0.3$ points, although Gemma-3-IT favors question repetition over image repositioning\rev{; question-first repetition (\TVTFmt{}) remains below \BaseFmt{} across evaluated families (Appendix~\ref{app:tvt})}. To examine how access to the question affects visual attention, we inspect cross-image attention in the language decoder.\par\endgroup
\end{minipage}\hfill
\begin{minipage}[t]{0.51\linewidth}
\vspace{0pt}
\makeatletter\def\@captype{table}\makeatother
\setlength{\abovecaptionskip}{0pt}
\centering
\setlength{\tabcolsep}{0.6pt}
\renewcommand{\arraystretch}{1.4}
\caption{Gains over \BaseFmt{} under default prompts (points), averaged over the 12 benchmarks. Family rows average over their models; the last row weights the three families equally.}
\label{tab:replay_variants}
\fontsize{8pt}{9.6pt}\selectfont
\begin{tabular*}{\linewidth}{@{\extracolsep{\fill}}lcccc@{}}
\toprule
\textbf{Family} & {\scriptsize\VVTFmt{}} & {\scriptsize\VTTFmt{}} & {\scriptsize\VTVFmt{}} & {\scriptsize\SLPFmt{}} \\
\midrule
Qwen2.5-VL & +0.4 & +0.1 & +0.8 & \textbf{+0.9} \\
Gemma-3-IT & +0.2 & +0.5 & +0.3 & \textbf{+0.8} \\
MiniCPM-V/o 4.5 & +0.5 & +0.3 & +0.4 & \textbf{+0.7} \\
\midrule
Average & +0.4 & +0.3 & +0.5 & \textbf{+0.8} \\
\bottomrule
\end{tabular*}
\end{minipage}
\par\vspace{4pt}

\textbf{Question-guided redirection of visual attention.}
Figure~\ref{fig:dynamath_sample} visualizes cross-image attention from $\Vis_2$ patch queries to $\Vis_1$ keys on \rev{Qwen2.5-VL-7B} across three representative tasks (\rev{layers 24--27; all-layer analyses for both 7B and 32B models and} protocol in Appendix~\ref{app:attention_protocol}). In all three cases, placing the image copy after the question shifts attention directly toward question-relevant regions (green boxes). Queried about a player's bib color (left), \VTVFmt{} shifts attention from the kicker and ball to that player; queried about structures above the root cap (middle) or the top of a food web (right), it shifts attention from diagram boundaries and producers to the queried regions. These examples show that access to the question can redirect second-image attention toward relevant evidence in the first image. Under the causal mask, \(\Vis_2\) has the same prefix in \VTVFmt{} and \SLPFmt{}, so these attention maps also hold for SLP.

\Needspace{6\baselineskip}
\subsubsection{What Role Does the Second Image Play?}
\label{sec:summarizer}

\hypobox{Finding 2}{Added visual tokens act as a prefill scratchpad, storing answer-relevant information for later tokens to read.}{They need no task-relevant pixels: even a blank image in its place can improve reasoning.}

To test whether the scratchpad's hidden states retain answer-relevant input information, we prefill \(\langle\Vis_1,\Txt_1,\Rblk,\text{``Answer:''}\rangle\), where the appended block \(\Rblk\) reads the image and question. We then predict the next option letter with attention restricted to \(\Rblk\) and the forced ``Answer:'' cue, itself blocked from directly attending to \(\Vis_1\) and \(\Txt_1\) (Figure~\ref{fig:h3_readout_probe}; Appendix~\ref{app:scratchpad}). We call this test the \emph{scratchpad probe}. The \emph{probe gain} is its accuracy gain over a \emph{cue-only baseline}, where only the cue is available for prediction. We test image and text blocks with matched core token counts, except for the question copy (Table~\ref{tab:h3_carriers}).

\textbf{Readout from the image copy recovers most of the gain.}
With \(\Rblk=\Vis_2\), the readout recovers 81\%--103\% of the gain of the unrestricted standard prompt (Full), despite having no direct access to \(\Vis_1\) or \(\Txt_1\) (Table~\ref{tab:h3_carriers}).

\begin{wrapfigure}{r}{0.54\linewidth}
\vspace{-0.6\baselineskip}
\centering
\setlength{\abovecaptionskip}{2pt}
\includegraphics[width=\linewidth]{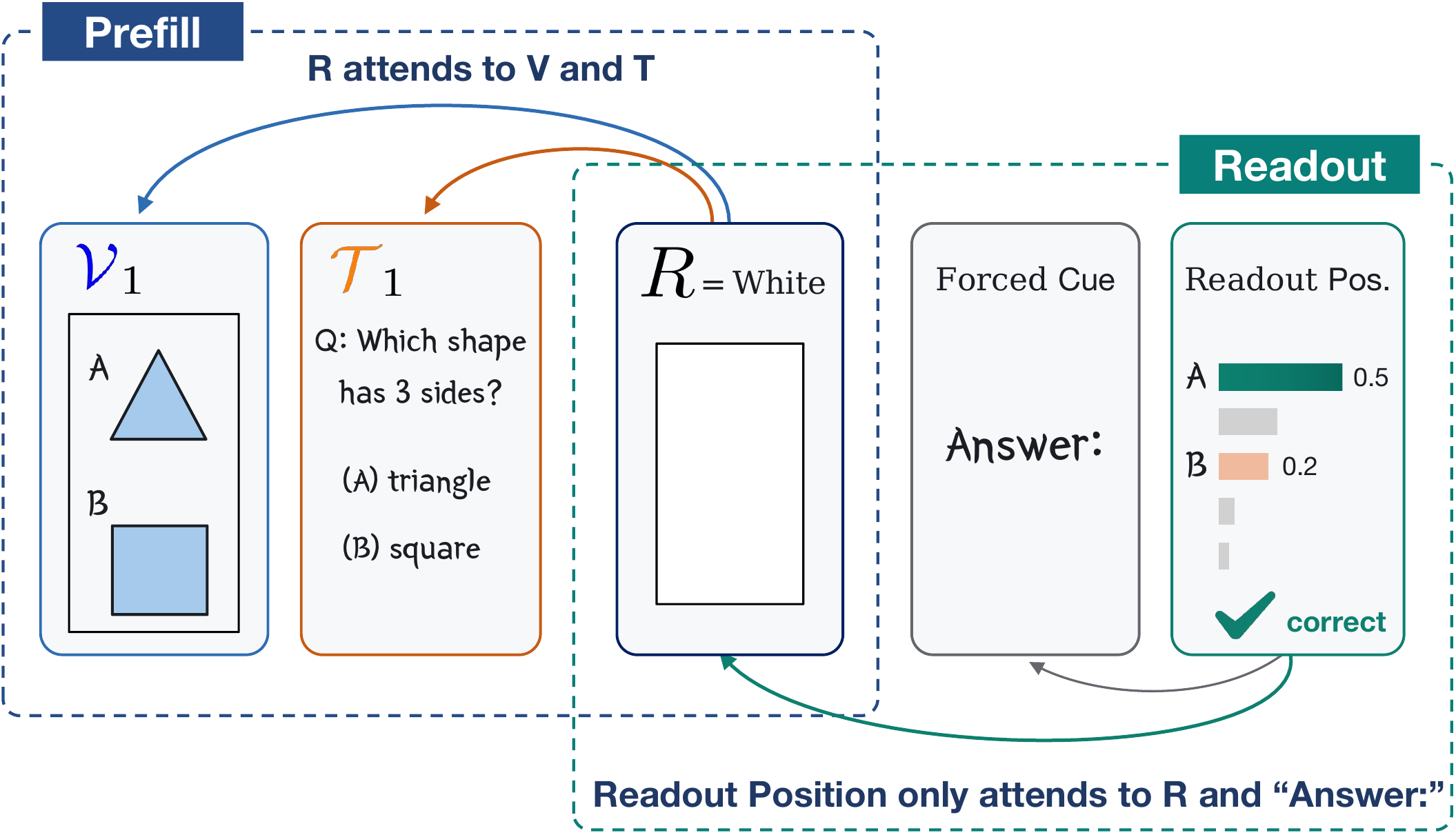}
\caption{Scratchpad probe with a white image as \(\Rblk\). Arrows point from queries to attended blocks.}
\label{fig:h3_readout_probe}
\end{wrapfigure}
\begingroup\hyphenpenalty=5000\exhyphenpenalty=5000
\textbf{Content-free blocks enable answering.}
Remarkably, \(\Rblk\) need not contain task information: content-free blocks achieve substantial probe gains across all tested models (up to $+35.8$ on MiniCPM-o 4.5 and $+40.9$ on Qwen2.5-VL-32B). Blocking \(\Rblk\)'s access to the earlier context during prefill largely removes these gains, returning accuracy to within 4.6 points of the cue-only baseline (Appendix~\ref{app:scratchpad}). The gains therefore primarily depend on input information written into \(\Rblk\), not on the block itself shifting the model's option preferences. The effective block varies by model: on Qwen2.5-VL, white and yellow images yield near-zero gains, whereas noise yields $+21.1$ to $+33.4$ points; on MiniCPM-V/o 4.5 and Gemma-3-12B-IT, uniform images yield strong gains. One possible explanation is that Qwen2.5-VL maps a uniform image to identical visual tokens that are hard to perform computation on, whereas the other models add absolute position embeddings that keep these tokens distinct (Appendix~\ref{app:scratchpad}).\par\endgroup

\begin{table}[htbp]
\centering
\footnotesize
\setlength{\abovecaptionskip}{2pt}
\setlength{\tabcolsep}{1.5pt}
\renewcommand{\arraystretch}{1.25}
\caption{Probe gains over the cue-only baseline. \emph{Full}: the standard prompt \(\langle\Vis_1,\Txt_1,\text{``Answer:''}\rangle\) with unrestricted attention. All blocks match \(\Vis\) in token count except Copy \(\Txt\); Lorem is fixed Lorem ipsum text. Boldface marks the best block, and underlining marks the best content-free block in each row.}
\label{tab:h3_carriers}
\begin{tabular*}{\linewidth}{@{\extracolsep{\fill}}lccccccccc@{}}
\toprule
\multirow{2}{*}[-0.6ex]{\textbf{Model}} & \multirow{2}{*}[-0.6ex]{\textbf{Full}} & \multicolumn{4}{c}{\textbf{\(\Rblk\): image block}} & \multicolumn{4}{c}{\textbf{\(\Rblk\): text block}} \\
\cmidrule(lr){3-6}\cmidrule(l){7-10}
& & {Copy \(\Vis\)} & {White} & {Yellow} & {Noise} & {Copy \(\Txt\)} & {Dots} & {Spaces} & {Lorem} \\
\midrule
{Qwen2.5-VL-3B} & +35.3 & \textbf{+28.7} & +1.6 & +0.0 & \underline{+28.0} & +13.1 & +0.8 & +11.3 & -0.2 \\
{Qwen2.5-VL-7B} & +39.4 & \textbf{+34.1} & +2.2 & -1.3 & \underline{+33.4} & +33.1 & -0.9 & +32.2 & +8.3 \\
{Qwen2.5-VL-32B} & +44.4 & +38.1 & -0.9 & -0.8 & +21.1 & +38.8 & +31.7 & \underline{\textbf{+40.9}} & -0.9 \\
\midrule
{MiniCPM-V 4.5} & +37.2 & +30.2 & \underline{\textbf{+32.0}} & +31.7 & +30.5 & +23.8 & +0.6 & +22.5 & -1.3 \\
{MiniCPM-o 4.5} & +39.4 & +34.5 & \underline{\textbf{+35.8}} & +33.6 & +23.9 & +20.0 & -0.8 & +0.2 & +0.5 \\
\midrule
{Gemma-3-12B-IT} & +33.8 & +34.8 & +34.8 & \underline{\textbf{+35.0}} & +33.2 & +30.9 & +31.1 & +29.2 & +30.8 \\
\bottomrule
\end{tabular*}
\end{table}

\textbf{Image blocks generally outperform text blocks.}
The best image block outperforms the best text block for five of six models, with an average margin of 7.0 points across all six (Table~\ref{tab:h3_carriers}); among content-free blocks alone, the best one is also an image block for the same five models. Qwen2.5-VL-32B is the only exception, for which spaces work best. The block's modality thus matters: the visual block is the more effective scratchpad.

\textbf{Even a white image can improve reasoning.}
The white image yields substantial probe gains on MiniCPM-o 4.5 and Gemma-3-12B-IT but not on Qwen2.5-VL-32B. We test whether this pattern carries over to standard generation by inserting a white image between the two question copies in \VTTFmt{}, giving \(\langle\Vis,\Txt,\square,\Txt\rangle\). Across the seven reasoning benchmarks, the white image improves scores over \VTTFmt{} by $+1.6$ points on MiniCPM-o 4.5 and $+0.9$ on Gemma-3-12B-IT, while Qwen2.5-VL-32B gains only $+0.1$ (Table~\ref{tab:h3_white}). This matching pattern supports the scratchpad interpretation: like filler tokens in language models~\citep{pfau2024dot}, a blank image block retains answer-relevant information without adding task-relevant pixels. \rev{At the same time, probe readability alone does not guarantee equal utilization during free generation: length-matched text fillers (\(\langle\Vis,\Txt,\text{Spaces},\Txt\rangle\)) yield much smaller reasoning gains than readable visual blocks, while attention-blocking and equal-input-budget order controls (\VVTTFmt{}) confirm that question-conditioned visual states in \(\Vis_2\) contribute to the \SLP{} improvement (Appendix~\ref{app:scratchpad_generation}).}

\begin{table}[!t]
\centering\footnotesize
\setlength{\abovecaptionskip}{0pt}
\sisetup{mode=text,retain-explicit-plus}
\newcommand{\tprobe}[1]{\tablenum[table-format=+2.1]{#1}}
\newcommand{\tdelta}[1]{\tablenum[table-format=+1.1]{#1}}
\caption{White-image gains. Probe gains are from Table~\ref{tab:h3_carriers}. The $\Delta$ columns compare \(\langle\Vis,\Txt,\square,\Txt\rangle\) with each prompt under standard generation. Underlining marks Qwen2.5-VL-32B's near-zero probe gain and small gain over \VTTFmt{}.}
\label{tab:h3_white}
\setlength{\tabcolsep}{11pt}
\renewcommand{\arraystretch}{1.12}
\begin{tabular}{@{}lccccccc}
\toprule
\multirow{2}{*}[-0.6ex]{\textbf{Model}} & \multirow{2}{*}[-0.6ex]{\textbf{Probe gain}} & \multicolumn{2}{c}{$\Delta$ vs.\ \VTTFmt{}} & \multicolumn{2}{c}{$\Delta$ vs.\ \SLPFmt{}} & \multicolumn{2}{c}{$\Delta$ vs.\ \BaseFmt{}} \\
\cmidrule(lr){3-4}\cmidrule(lr){5-6}\cmidrule(lr){7-8}
& & Reas. & Perc. & Reas. & Perc. & Reas. & Perc. \\
\midrule
Qwen2.5-VL-32B & \underline{\tprobe{-0.9}} & \underline{\tdelta{+0.1}} & \underline{\tdelta{+0.2}} & \tdelta{-1.9} & \tdelta{+0.1} & \tdelta{+0.2} & \tdelta{+0.5} \\
MiniCPM-o 4.5  & \tprobe{+35.8} & \tdelta{+1.6} & \tdelta{+0.1} & \tdelta{+1.1} & \tdelta{-0.2} & \tdelta{+2.0} & \tdelta{+0.4} \\
Gemma-3-12B-IT & \tprobe{+34.8} & \tdelta{+0.9} & \tdelta{-0.4} & \tdelta{-0.3} & \tdelta{0.0} & \tdelta{+1.1} & \tdelta{+1.3} \\
\bottomrule
\end{tabular}
\end{table}

\textbf{Breaking exact spatial correspondence further improves SLP.}
The image copy is not always the best block: a content-free block beats it in the scratchpad probe for four of six models (Table~\ref{tab:h3_carriers}), and a white image beats it in reasoning with MiniCPM-o 4.5 under standard generation (Table~\ref{tab:h3_white}). We hypothesize that exact correspondence draws \(\Vis_2\)'s attention toward \(\Vis_1\) at the expense of attention to the question and computation within \(\Vis_2\). We therefore cyclically shift \(\Vis_2\) by 7--28 pixels. Averaged across shifts and directions, reasoning scores increase over intact SLP by $+0.3$, $+0.5$, and $+1.3$ points for Qwen2.5-VL-32B, Qwen2.5-VL-3B, and MiniCPM-o 4.5. On 100 LogicVista prompts, an attention probe on Qwen2.5-VL-32B shows that rightward shifts move \(\Vis_2\)'s attention away from \(\Vis_1\) toward the question and \(\Vis_2\) itself (Appendix~\ref{app:shifted_copy}).

Findings~1 and 2 identify two complementary functions of the added image block: its tokens attend to \(\Vis_1\) under guidance from the question, and its hidden states retain answer-relevant input information that the model reads when answering. The two functions are linked: the scratchpad states are written by reading the image and question. In SLP, \(\Txt_2\) is the first block that can read this scratchpad during prefill, processing the question again with access to the question-guided \(\Vis_2\) and completing unified multimodal encoding. \VTVFmt{} lacks this step: appending \(\Txt_2\) raises the average gain from $+0.5$ to $+0.8$, and SLP yields the largest gain in each model family (Table~\ref{tab:replay_variants}). The replayed segments thus support contextual computation as well as another view of the same pixels.

\subsubsection{Why Does Reasoning Not Resolve Partial Observation?}
\label{sec:cot_interaction}

\providecommand{\Q}{\ensuremath{\langle \Txt \rangle}}
\providecommand{\QQ}{\ensuremath{\langle \Txt, \Txt \rangle}}

\hypobox{Finding 3}{Reasoning complements rather than resolves cross-modal partial observation.}{Generated text cannot reorganize visual states formed during prefill; SLP remains beneficial across tested reasoning budgets.}

\begin{wraptable}{r}{0.57\linewidth}
\vspace{-0.1\baselineskip}
\setlength{\abovecaptionskip}{0pt}
\caption{Repetition gains (points) on 32B models. $\Delta = \mathrm{Default}-\mathrm{Direct}$ is the difference between gains.}
\label{tab:cot_replay_interaction}
\centering\small
\setlength{\tabcolsep}{3pt}
\renewcommand{\arraystretch}{1.2}
\begin{tabular*}{\linewidth}{@{\extracolsep{\fill}}llccc@{}}
\toprule
\textbf{Input} & \textbf{Model} & Direct & Default & $\Delta$ \\
\midrule
\QQ{} & {Qwen2.5-32B} & +3.1 & +0.7 & -2.4 \\
\SLPFmt{} & {Qwen2.5-VL-32B} & +1.3 & +1.4 & \textbf{+0.0} \\
\bottomrule
\end{tabular*}
\end{wraptable}
\begingroup\hyphenpenalty=5000\exhyphenpenalty=5000
Findings~1 and 2 show that the second look reorganizes visual information under question guidance and provides a prefill scratchpad. We next ask whether generated reasoning can replace these functions. In text-only models, reasoning may repeat the question and reduce the benefit of prompt repetition, as suggested by prior work~\citep{leviathan2025promptrepetitionimprovesnonreasoning}. We compare direct-answer instructions with default benchmark prompts, which permit reasoning without requiring it. The text-only experiment compares \Q{} and \QQ{} on Qwen2.5-32B over five text benchmarks; the VLM experiment compares \BaseFmt{} and \SLPFmt{} on Qwen2.5-VL-32B over the 12 main benchmarks (per-benchmark results in Appendix~\ref{app:text_replay}).\par\endgroup

\textbf{The text-only repetition gain is smaller with default prompts.}
Repeating the question improves the average score by $3.1$ points with direct-answer instructions and $0.7$ with default prompts, a reduction of $2.4$ points (Table~\ref{tab:cot_replay_interaction}).

\textbf{Generated reasoning does not replace prefill replay.}
SLP improves the VLM average by $1.3$ points with direct-answer instructions and $1.4$ with default prompts (Table~\ref{tab:cot_replay_interaction}). The similar gains indicate that generated reasoning does not substitute for the unified encoding of visual and textual information established by replaying the pair during prefill \rev{in these settings; explicit reasoning-effort controls on Qwen3.8-27B (\texttt{low}, \texttt{medium}, and \texttt{xhigh}, up to $>4{,}200$ output tokens) likewise retain consistent \SLP{} gains of $+2.2$ to $+2.9$ points (Appendix~\ref{app:new_models})}. From the modality perspective, reasoning produces text, which may be a less effective scratchpad than visual tokens. This is consistent with the generally stronger image-block results in Finding~2, though that probe compares fixed input blocks rather than generated rationales (Section~\ref{sec:summarizer}). From the functional perspective, the next-token prediction objective alone may not be sufficient to train generated rationales to integrate multimodal input in the same way as prefill replay. Together, these differences suggest that input-side replay \rev{complements generation-time reasoning across tested reasoning budgets}.

\providecommand{\SLPName}{Second-Look Prompting}
\providecommand{\SLP}{SLP}

\subsection{Learning to Use the Second Look}
\label{subsec:training-study}

SLP leverages an existing input capability without adaptation. We next test whether continued training amplifies this latent capability. Starting from LLaVA-OneVision-7B-SI~\citep{li2024llavaonevision}, we conduct two training runs on the same single-turn, single-image instruction subset, one using \BaseFmt{} (Standard) and the other \SLPFmt{} (SLP), with optimizer steps and hyperparameters matched (Appendix~\ref{app:training_study}).

\begin{figure}[h]
\centering
\begin{minipage}[t]{0.63\linewidth}
\vspace{0pt}
\centering
\includegraphics[width=\linewidth]{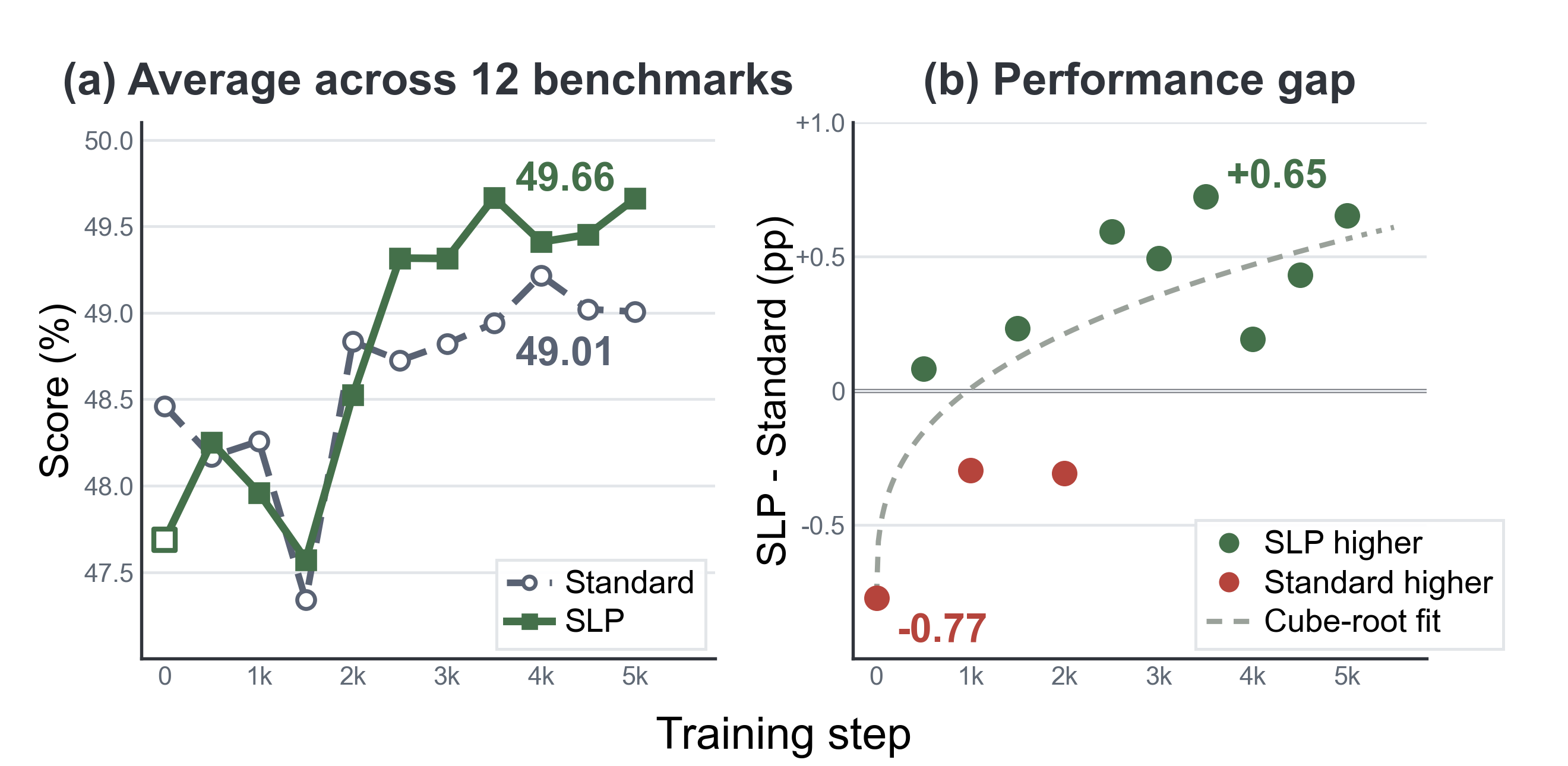}
\end{minipage}\hfill
\begin{minipage}[t]{0.36\linewidth}
\vspace{0pt}
\centering\small
\makeatletter\def\@captype{table}\makeatother
\caption{Scores for each training and inference prompt combination, averaged over the same 12 benchmarks.}
\label{tab:training_cross_format}
\setlength{\tabcolsep}{4pt}
\renewcommand{\arraystretch}{1.2}
\begin{tabular*}{\linewidth}{@{\extracolsep{\fill}}lcc@{}}
\toprule
& \multicolumn{2}{c}{\textbf{Inference}} \\
\cmidrule(lr){2-3}
\textbf{Training} & Standard & SLP \\
\midrule
Standard & 49.01 & 48.41 \\
SLP & 49.00 & \textbf{49.66} \\
\bottomrule
\end{tabular*}
\end{minipage}
\caption{Continued training on \SLPFmt{} vs.\ a matched \BaseFmt{} control. (a) Overall average over 12 benchmarks, with each run evaluated using its training prompt; step 0 uses the shared LLaVA-OneVision-7B-SI checkpoint. (b) Per-checkpoint SLP$-$Standard gap with a descriptive cube-root fit; the dotted segment is extrapolated beyond the observed checkpoints.}
\label{fig:training_study_single_column}
\end{figure}

\textbf{Training unlocks the second-look gain.}
At initialization, the single-image checkpoint has never seen interleaved inputs, scoring 47.69 under SLP versus 48.46 under Standard. Under continued training, it learns to utilize the repeated presentation: SLP gains $+1.97$ points versus $+0.55$ for Standard, establishing a $+0.65$-point lead overall (Figure~\ref{fig:training_study_single_column}). The gap widens on both suites, from $-0.03$ to $+0.94$ points on reasoning and from $-1.81$ to $+0.25$ on perception.

\textbf{SLP training preserves the standard-prompt average.}
At step 5,000, the SLP-trained and Standard-trained checkpoints achieve nearly identical macro averages under \BaseFmt{} inference (49.00 vs.\ 49.01 across 12 benchmarks; Table~\ref{tab:training_cross_format}). Individual benchmark scores change in both directions (Appendix~\ref{app:training_study}). Under \SLPFmt{} inference, the SLP-trained checkpoint outperforms the Standard-trained checkpoint by $1.25$ points on the same 12-benchmark macro average.

\Needspace{12\baselineskip}
\subsection{Inference Efficiency}
\label{sec:efficiency}

\begin{wraptable}{r}{0.43\linewidth}
\vspace{-0.4\baselineskip}
\setlength{\abovecaptionskip}{0pt}
\caption{Mean end-to-end latency (s), averaged within families. \(\Delta\) is SLP minus Standard.}
\label{tab:efficiency}
\centering
\footnotesize
\setlength{\tabcolsep}{1.5pt}
\renewcommand{\arraystretch}{1.2}
\begin{tabular*}{\linewidth}{@{\extracolsep{\fill}}l c c c@{}}
\toprule
\textbf{Family}
& \textbf{\BaseFmt{}}
& \textbf{\SLPFmt{}}
& \textbf{\(\Delta\)} \\
\midrule
Qwen2.5-VL & 3.12 & 3.28 & $+0.16$ \\
Gemma-3-IT & 2.32 & 2.16 & $-0.16$ \\
MiniCPM-V/o 4.5 & 4.81 & 4.84 & $+0.03$ \\
\bottomrule
\end{tabular*}
\end{wraptable}
We measure end-to-end (E2E) latency on all nine open-source models using default prompts and a shared 10\% subsample of each benchmark (Appendix~\ref{app:efficiency}). Family-average changes are $+0.16$ seconds for Qwen2.5-VL, $-0.16$ seconds for Gemma-3-IT, and $+0.03$ seconds for MiniCPM-V/o 4.5 (Table~\ref{tab:efficiency}). Shorter generations may offset the extra prefill cost: mean output length in characters decreases for seven of nine models (Appendix~\ref{app:length}), also indicating that SLP's gains do not rely on longer generations. \rev{Moreover, downsampling only \(\Vis_2\) to half or a quarter of its visual-token budget further lowers latency while slightly improving accuracy over full-resolution \SLP{} (Appendix~\ref{app:lowres}).}

\section{Related Work}
\label{sec:related}

\textbf{Prompt repetition and reasoning strategies.}
Reasoning strategies include CoT~\citep{wei2022chain,kojima2022large}, self-consistency~\citep{wang2022self}, and structured search~\citep{yao2023tree,besta2024graph}. Other work studies sensitivity to input order~\citep{liu2024lost,chen2024premise} and improves reasoning, embeddings, or sequence labeling by repeating the input~\citep{xu2024re,springer2025repetition,kukic2026sequence}. Repetition gains vary by task~\citep{shaier2024asking} and can diminish when reasoning is permitted~\citep{leviathan2025promptrepetitionimprovesnonreasoning}. We show that repeating an image--question pair establishes unified multimodal encoding, with gains that persist when reasoning is permitted.

\textbf{Input order and cross-modal conditioning.}
Input order affects VLMs~\citep{tan2024order,wardle2024imagefirst,wei2026multimodalorder}. Architectural designs enable question-conditioned visual features~\citep{dai2023instructblip,ganz2024question}. Concurrent work studies \SLPFmt{} for visual steering and question access in natural-image VQA and compositional matching~\citep{abhinandan2026asktwice}. However, transfer to broader reasoning tasks and frontier proprietary VLMs remains untested. We evaluate SLP across 12 reasoning and perception benchmarks and 12 open-source and proprietary VLMs, and investigate how visual replay supports computation beyond re-exposure to the image.

\textbf{Reasoning with visual intermediates.}
Thinking-with-images methods interactively extract, crop, mark, or sketch images during reasoning~\citep{su2025thinking,gao2025interleaved,zhou2024image,hu2024visual,liu2024chain,cao2024dualfocus,zheng2026deepeyes}; multimodal CoT produces rationales conditioned on both modalities~\citep{zhang2023multimodal,mitra2024compositional,zhao2024bba}; and VGR replays detected image regions during reasoning~\citep{wang2026vgr}. Unlike these generation-time methods, SLP operates on the input side.

\textbf{Computation with additional tokens.}
Pause tokens~\citep{goyal2024pause}, filler tokens~\citep{pfau2024dot}, and vision-transformer registers~\citep{darcet2024registers} show how additional token positions can support computation, typically with training. SLP instead repeats the image--question pair to enable question-conditioned visual processing during prefill. Our analyses further show that the added visual tokens can serve as a scratchpad without additional training (Section~\ref{sec:summarizer}).

\section{Limitations}
\label{sec:limitations}

SLP is a simple input-side intervention, and its scope and costs have three main limitations. First, duplicate images may interfere with position-sensitive tasks: in stress tests, counting and spatial relations hold ($-0.12$ on CountQA; $+0.54$ on SpatialMQA), but grounding drops by $1.11$ points on RefCOCOg (Appendix~\ref{app:duplicate_image_stress}). Second, training on SLP takes $2.25\times$ the wall-clock time of standard training due to the duplicated prefix (32.0 vs.\ 14.2 hours for 5{,}000 steps on 16 A100 GPUs; Appendix~\ref{app:training_study}). Third, SLP does not yet offer a unified treatment of multi-image or interleaved image--text inputs: our experiments are limited to prompts whose original form is \BaseFmt{}, and how to replay other input organizations is left for future work.

\section{Conclusion}

We introduce Second-Look Prompting (SLP), a training-free input presentation that addresses the cross-modal partial-observation problem by establishing unified multimodal encoding under causal attention. SLP reshapes prefill computation: the repeated image revisits the earlier visual content under the question's guidance, and the repeated question then conditions on the resulting visual states. These added visual tokens also act as a prefill scratchpad, retaining answer-relevant input information, so that even a blank second look can aid reasoning. SLP's gains persist when reasoning is allowed, unlike those of text-only repetition. Continued training improves performance under SLP while retaining a similar average under standard prompting. These results position input-side prompting as an accessible control surface complementary to generation-time reasoning. They also suggest that the potential of input presentation remains underexplored and raise a broader question: what properties distinguish visual from textual tokens during prefill, and how do these differences shape the integration of the image and question?

\label{sec:main-text-end}

\clearpage
\setlength{\textfloatsep}{\savedtextfloatsep}
\subsection*{AI use statement}
We used generative AI tools to aid and polish writing (language editing,
manuscript organization, terminology and repetition checks, and LaTeX formatting)
and for retrieval and discovery (finding related work and verifying bibliographic
entries). We have reviewed all AI-assisted content and take full responsibility
for the final content of this paper.

\subsection*{Ethics statement}
This paper presents work whose goal is to advance the field of machine
learning. There are many potential societal consequences of our work,
none of which we feel must be specifically highlighted here.

\subsection*{Reproducibility statement}
Prompt construction is described in Section~\ref{sec:prelim}. Prompt templates,
decoding settings, and benchmark sampling are listed in
Appendix~\ref{app:prompts_decoding}. Protocols \rev{and additional controls} for the attention analyses, \rev{order baselines,} and scratchpad probe are given
in Appendices~\ref{app:order_reversal}, \ref{app:attention_protocol}, and~\ref{app:scratchpad}; the
continued-training settings and results appear in
Appendix~\ref{app:training_study}\rev{, low-resolution replay and newer VLM evaluations appear in Appendices~\ref{app:lowres} and~\ref{app:new_models}, and our implementation and evaluation configurations are available at \url{https://github.com/zycccishere/second-look-prompting}}.

\bibliography{main}
\bibliographystyle{iclr2027_conference}

\clearpage
\appendix
\raggedbottom
\let\appendixsection\section
\renewcommand{\section}{\FloatBarrier\appendixsection}
\sisetup{mode=text,retain-explicit-plus,detect-weight}
\Needspace{10\baselineskip}
\section{Prompt and Decoding Details}
\label{app:prompts_decoding}

\subsection{Prompt Templates}
\label{app:templates}

\textbf{Default prompts.}
All main experiments use each benchmark's official prompt verbatim, including the question, any benchmark-provided hint, the option list, and the official answer-format instruction. A schematic multiple-choice prompt is

\begin{quote}
\small
\begin{verbatim}
Question: {question}
Options:
A. {A}
B. {B}
...
Please select the correct answer from the options above.
\end{verbatim}
\end{quote}

Free-form benchmarks pose the question with their official answer-format hint (e.g., DynaMath's JSON output with \texttt{solution} and \texttt{short answer} keys). These prompts permit but do not require intermediate reasoning. The same prompt is used for all models and input presentations: a presentation such as \VTVFmt{} or \SLPFmt{} only reorders or duplicates the image and question segments verbatim. Each model uses its official chat template and native image tokens.

\textbf{Direct-answer prompts (Section~\ref{sec:cot_interaction}).}
For the direct-answer ablation, we append an instruction that asks for the final answer without intermediate reasoning (Table~\ref{tab:app_templates}); for DynaMath, it replaces the official JSON-format instruction.

\begin{table}[htbp]
\centering
\footnotesize
\renewcommand{\arraystretch}{1.15}
\caption{Direct-answer instructions, appended to the default prompt. A few benchmarks keep their official phrasing (e.g., VisualPuzzles keeps ``the option letter from the given choices''; We-Math embeds the instruction in its official hint).}
\label{tab:app_templates}
\begin{tabularx}{\linewidth}{@{}>{\raggedright\arraybackslash}p{0.12\linewidth} >{\raggedright\arraybackslash}p{0.24\linewidth} >{\raggedright\arraybackslash}X@{}}
\toprule
\textbf{Answer type} & \textbf{Benchmarks} & \textbf{Instruction} \\
\midrule
Option letter
& LogicVista, VisualPuzzles, MMMU, We-Math, AI2D, SEED-Bench-2-Plus, MMBench, MMStar
& \texttt{Return only the final option letter. Do not write 'Answer:'. Do not include explanation, reasoning, derivation, punctuation, or extra text. Output only one option letter, such as A, B, C, D, E, or F.} \\
\addlinespace
Free-form
& MATH-Vision, MathVista
& \texttt{Return only the final answer. Do not write 'Answer:'. Do not include explanation, reasoning, derivation, units unless required by the question, or extra text.} \\
\addlinespace
Recognized text
& OCRBench
& \texttt{Return only the recognized text. Do not write 'Answer:'. Do not include explanation, reasoning, punctuation not present in the image, or extra text.} \\
\addlinespace
Number
& DynaMath
& \texttt{Return only the final numeric answer as a three-digit floating-point number. Do not output JSON. Do not include explanation, reasoning, derivation, units, or extra text.} \\
\bottomrule
\end{tabularx}
\end{table}

\subsection{Decoding Configurations}
\label{app:decoding}

Table~\ref{tab:app_decoding} lists the decoding configurations. Within each comparison, all input presentations share the same configuration. Open-source models are served with vLLM, and proprietary models are queried through OpenAI-compatible chat-completion endpoints.

\begin{table}[htbp]
\centering
\footnotesize
\renewcommand{\arraystretch}{1.15}
\caption{Decoding configurations. Unlisted sampling parameters keep engine or provider defaults. Qwen2.5-VL on LogicVista uses near-greedy sampling (temperature 0.01, top-$p$ 1.0, repetition penalty 1.05). On the perception suite, MiniCPM-V uses regular CoT throughout; MiniCPM-o follows the vendor defaults, using no thinking on AI2D, OCRBench, and SEED-Bench-2-Plus and regular CoT on MMBench and MMStar.}
\label{tab:app_decoding}
\begin{tabular*}{\linewidth}{@{\extracolsep{\fill}}llccl@{}}
\toprule
\textbf{Model} & \textbf{Decoding} & \textbf{Temp.} & \textbf{Max tokens} & \textbf{Reasoning control} \\
\midrule
Qwen2.5-VL (3B--72B) & greedy & 0 & 2,048 & none \\
Gemma-3-IT (4B--27B) & greedy & 0 & 4,096 & none \\
MiniCPM-V/o 4.5, reasoning & top-$p$ 1.0 & 0.7 & 16,384 & long-CoT thinking \\
MiniCPM-V/o 4.5, perception & greedy, rep.\ pen.\ 1.2 & 0 & 16,384 & per benchmark \\
\midrule
GPT-5.4 mini & API & 0 & 32,768 & reasoning effort: minimal \\
Gemini 3 Flash & API & 0 & 32,768 & non-thinking \\
Claude Haiku 4.5 & API & 0 & 32,768 & provider default \\
\bottomrule
\end{tabular*}
\end{table}

\subsection{Benchmark Splits and Subsampling}
\label{app:benchmarks}

Open-source models are evaluated on the full splits. Due to API cost, proprietary models are evaluated on a deterministic subset of at most 1{,}000 examples per benchmark, drawn with the benchmark-specific seed \texttt{int(sha256("\{seed\}:\{dataset\}")[:16], 16)} and global seed 20260627. LogicVista, MathVista \texttt{testmini}, and OCRBench have at most 1{,}000 examples and are used in full.

\section{Confidence Intervals for the Main Results}
\label{app:significance}

Table~\ref{tab:significance} gives a 95\% confidence interval for each model's overall gain in Table~\ref{tab:slp-format-total}. Because \BaseFmt{} and \SLPFmt{} are evaluated on the same benchmarks with identical decoding, we treat the benchmark as the paired unit: we resample the 12 per-benchmark gains with replacement 20{,}000 times and report the percentile interval of the mean. The interval thus reflects how consistently the gain holds across benchmarks. For reference, within-benchmark sampling contributes a standard error of about 0.5 points to each overall average under a binomial model.

\begin{table}[htbp]
\centering
\small
\setlength{\tabcolsep}{8pt}
\renewcommand{\arraystretch}{1.1}
\caption{Overall gain of \SLPFmt{} over \BaseFmt{} (12-benchmark average) with 95\% paired-bootstrap confidence intervals. Improved/declined counts benchmarks with positive/negative gains; the rest are ties at the reported precision.}
\label{tab:significance}
\begin{tabular}{@{}lS[table-format=+1.2]cc@{}}
\toprule
\textbf{Model} & {$\boldsymbol{\Delta}$} & \textbf{95\% CI} & \textbf{Improved / declined} \\
\midrule
GPT-5.4 mini     & +1.68 & [+0.80, +2.68] & 10 / 1 \\
Gemini 3 Flash   & +1.08 & [+0.21, +2.15] & 8 / 4 \\
Claude Haiku 4.5 & +1.25 & [+0.83, +1.72] & 12 / 0 \\
\midrule
Qwen2.5-VL-72B & +1.26 & [+0.59, +1.92] & 10 / 1 \\
Qwen2.5-VL-32B & +1.36 & [+0.73, +1.99] & 11 / 1 \\
Qwen2.5-VL-7B  & +0.10 & [\textminus 0.53, +0.71] & 5 / 5 \\
Qwen2.5-VL-3B  & +0.78 & [+0.03, +1.62] & 7 / 4 \\
\midrule
MiniCPM-o 4.5 & +0.76 & [+0.18, +1.37] & 10 / 2 \\
MiniCPM-V 4.5 & +0.63 & [+0.07, +1.17] & 11 / 1 \\
\midrule
Gemma-3-27B-IT & +0.89 & [\textminus 0.10, +1.87] & 9 / 3 \\
Gemma-3-12B-IT & +1.33 & [+0.26, +2.62] & 10 / 2 \\
Gemma-3-4B-IT  & +0.22 & [\textminus 1.07, +1.47] & 6 / 5 \\
\bottomrule
\end{tabular}
\end{table}

\section{Order Reversal: Per-Benchmark Results}
\label{app:order_reversal}

Table~\ref{tab:app_order_reversal} gives the per-benchmark results behind Figure~\ref{fig:teaser} (left). Reversing \BaseFmt{} to \TVFmt{} lowers the overall average of every Qwen2.5-VL and Gemma-3-IT model by 2.1--6.3 points. The drop is larger on perception (3.9--12.4 points) than on reasoning (0.2--2.9 points).

\begin{table}[htbp]
\centering
\footnotesize
\setlength{\tabcolsep}{3pt}
\renewcommand{\arraystretch}{1.08}
\caption{Order reversal. Indented rows evaluate the model above with \TVFmt{} instead of \BaseFmt{}; deltas on the averages are relative to \BaseFmt{}.}
\label{tab:app_order_reversal}
\begin{tabular*}{\linewidth}{@{\extracolsep{\fill}}lcccccccc@{}}
\toprule
\multicolumn{9}{c}{\textbf{(a) Reasoning}} \\
\midrule
\textbf{Model} & \shortstack{Dyna\\Math} & \shortstack{MATH-\\Vision} & \shortstack{Logic\\Vista} & \shortstack{Visual\\Puzzles} & MMMU & MathVista & We-Math & \textbf{Avg.} \\
\midrule
{\footnotesize Qwen2.5-VL-72B} & 66.7 & 38.6 & 57.5 & 41.6 & 67.7 & 73.5 & 48.8 & 56.3 \\
{\footnotesize \hspace{0.8em}\textbf{\TVFmt{}}} & 66.4 & 39.3 & 57.7 & 44.5 & 66.3 & 70.6 & 48.5 & \scorered{56.2}{-0.2} \\
{\footnotesize Qwen2.5-VL-32B} & 67.0 & 37.3 & 54.4 & 38.2 & 65.8 & 72.7 & 46.9 & 54.6 \\
{\footnotesize \hspace{0.8em}\textbf{\TVFmt{}}} & 66.8 & 37.9 & 56.8 & 37.2 & 63.4 & 71.8 & 43.0 & \scorered{53.8}{-0.8} \\
{\footnotesize Qwen2.5-VL-7B} & 54.2 & 25.4 & 46.5 & 32.9 & 53.8 & 66.6 & 35.3 & 45.0 \\
{\footnotesize \hspace{0.8em}\textbf{\TVFmt{}}} & 53.6 & 27.7 & 45.4 & 32.6 & 49.5 & 64.9 & 31.9 & \scorered{43.7}{-1.3} \\
{\footnotesize Qwen2.5-VL-3B} & 43.6 & 21.7 & 36.0 & 28.5 & 48.4 & 62.1 & 21.6 & 37.4 \\
{\footnotesize \hspace{0.8em}\textbf{\TVFmt{}}} & 40.1 & 22.9 & 37.8 & 28.2 & 47.7 & 50.2 & 21.9 & \scorered{35.6}{-1.9} \\
\midrule
{\footnotesize Gemma-3-27B-IT} & 50.6 & 36.7 & 47.2 & 15.2 & 57.6 & 60.1 & 42.0 & 44.2 \\
{\footnotesize \hspace{0.8em}\textbf{\TVFmt{}}} & 42.8 & 38.2 & 38.3 & 17.2 & 54.5 & 58.5 & 40.2 & \scorered{41.4}{-2.9} \\
{\footnotesize Gemma-3-12B-IT} & 30.8 & 31.2 & 36.9 & 16.4 & 51.6 & 54.1 & 35.3 & 36.6 \\
{\footnotesize \hspace{0.8em}\textbf{\TVFmt{}}} & 33.0 & 32.5 & 34.5 & 17.6 & 50.2 & 51.6 & 26.7 & \scorered{35.2}{-1.5} \\
{\footnotesize Gemma-3-4B-IT} & 25.9 & 22.2 & 30.7 & 13.8 & 42.0 & 43.8 & 25.7 & 29.1 \\
{\footnotesize \hspace{0.8em}\textbf{\TVFmt{}}} & 23.5 & 22.4 & 26.9 & 13.8 & 44.1 & 42.5 & 27.5 & \scorered{28.7}{-0.5} \\
\bottomrule
\end{tabular*}
\par\medskip
\begin{tabular*}{\linewidth}{@{\extracolsep{\fill}}lccccccc@{}}
\toprule
\multicolumn{8}{c}{\textbf{(b) Perception and overall}} \\
\midrule
\textbf{Model} & AI2D & OCRBench & \shortstack{SEED-Bench-\\2-Plus} & MMBench & MMStar & \textbf{Avg.} & \textbf{Overall} \\
\midrule
{\footnotesize Qwen2.5-VL-72B} & 88.6 & 88.6 & 73.7 & 88.2 & 71.0 & 82.0 & 67.0 \\
{\footnotesize \hspace{0.8em}\textbf{\TVFmt{}}} & 82.5 & 87.8 & 69.8 & 82.6 & 61.3 & \scorered{76.8}{-5.2} & \scorered{64.8}{-2.3} \\
{\footnotesize Qwen2.5-VL-32B} & 83.6 & 85.5 & 72.6 & 85.5 & 66.8 & 78.8 & 64.7 \\
{\footnotesize \hspace{0.8em}\textbf{\TVFmt{}}} & 80.7 & 81.8 & 70.6 & 79.3 & 62.1 & \scorered{74.9}{-3.9} & \scorered{62.6}{-2.1} \\
{\footnotesize Qwen2.5-VL-7B} & 85.1 & 88.3 & 70.9 & 83.2 & 63.9 & 78.3 & 58.8 \\
{\footnotesize \hspace{0.8em}\textbf{\TVFmt{}}} & 74.0 & 85.6 & 62.2 & 71.5 & 54.8 & \scorered{69.6}{-8.7} & \scorered{54.5}{-4.4} \\
{\footnotesize Qwen2.5-VL-3B} & 81.5 & 82.3 & 68.5 & 77.1 & 55.7 & 73.0 & 52.3 \\
{\footnotesize \hspace{0.8em}\textbf{\TVFmt{}}} & 66.6 & 77.9 & 56.1 & 61.5 & 41.0 & \scorered{60.6}{-12.4} & \scorered{46.0}{-6.3} \\
\midrule
{\footnotesize Gemma-3-27B-IT} & 83.3 & 75.3 & 70.1 & 79.5 & 56.5 & 72.9 & 56.2 \\
{\footnotesize \hspace{0.8em}\textbf{\TVFmt{}}} & 76.0 & 73.7 & 64.6 & 72.1 & 47.9 & \scorered{66.8}{-6.1} & \scorered{52.0}{-4.2} \\
{\footnotesize Gemma-3-12B-IT} & 80.3 & 72.9 & 67.2 & 75.2 & 54.7 & 70.1 & 50.5 \\
{\footnotesize \hspace{0.8em}\textbf{\TVFmt{}}} & 74.8 & 70.1 & 62.4 & 73.5 & 47.8 & \scorered{65.7}{-4.3} & \scorered{47.9}{-2.6} \\
{\footnotesize Gemma-3-4B-IT} & 73.5 & 67.9 & 61.5 & 66.1 & 45.9 & 63.0 & 43.2 \\
{\footnotesize \hspace{0.8em}\textbf{\TVFmt{}}} & 64.1 & 63.9 & 55.3 & 63.2 & 41.6 & \scorered{57.6}{-5.4} & \scorered{40.7}{-2.5} \\
\bottomrule
\end{tabular*}
\end{table}

\subsection{\rev{Question--Image--Question Baseline}}
\label{app:tvt}

\rev{To test whether repeating only the question around a single image resolves question-first sensitivity, we evaluate \TVTFmt{} across all 12 benchmarks on Qwen2.5-VL-32B, MiniCPM-o 4.5, and Gemma-3-12B-IT (Table~\ref{tab:app_tvt}). Across all three models, \SLPFmt{} outperforms \TVTFmt{} by $+1.13$ to $+5.06$ points overall, whereas \TVTFmt{} remains below single-pass \BaseFmt{} ($63.96$ vs.\ $64.7$, $62.47$ vs.\ $62.8$, and $46.84$ vs.\ $50.5$ in Table~\ref{tab:slp-format-total}). On a shared 10\% subsample across all 12 benchmarks (2{,}610 examples on Qwen2.5-VL-32B; $4\times$A100, TP=4, batch size 1), \SLPFmt{} uses more input tokens than \TVTFmt{} ($3{,}014.1$ vs.\ $1{,}603.9$) yet generates 12.4\% fewer output tokens ($338.8$ vs.\ $386.6$) and finishes 6.6\% faster end to end ($6.32$ vs.\ $6.77$~s).}

\begin{table}[!t]
\centering
\footnotesize
\renewcommand{\arraystretch}{1.08}
\caption{\rev{Question--image--question (\TVTFmt{}) vs.\ \SLPFmt{} (Table~\ref{tab:slp-format-total}) across all 12 benchmarks.}}
\label{tab:app_tvt}
\filltabular{8}{@{}l*{8}{c}@{}}{%
\toprule
\multicolumn{9}{c}{\textbf{(a) Reasoning}} \\
\midrule
\textbf{Model} & \shortstack{Dyna\\Math} & \shortstack{MATH-\\Vision} & \shortstack{Logic\\Vista} & \shortstack{Visual\\Puzzles} & MMMU & MathVista & We-Math & \textbf{Avg.} \\
\midrule
Qwen2.5-VL-32B (\TVTFmt{}) & 67.05 & 39.44 & 56.38 & 40.92 & 63.01 & 74.20 & 49.43 & 55.78 \\
\rowcolor{colorhead}\hspace{0.8em}\textbf{\SLPFmt{}} & 67.5 & 38.8 & 57.7 & 40.8 & 67.7 & 74.9 & 49.4 & \textbf{56.7} \\
\midrule
MiniCPM-o 4.5 (\TVTFmt{}) & 61.06 & 45.36 & 50.78 & 19.18 & 68.61 & 71.70 & 57.05 & 53.39 \\
\rowcolor{colorhead}\hspace{0.8em}\textbf{\SLPFmt{}} & 62.5 & 46.1 & 54.1 & 15.3 & 70.8 & 74.2 & 57.0 & \textbf{54.3} \\
\midrule
Gemma-3-12B-IT (\TVTFmt{}) & 31.06 & 30.23 & 33.11 & 15.50 & 47.14 & 48.90 & 25.71 & 33.09 \\
\rowcolor{colorhead}\hspace{0.8em}\textbf{\SLPFmt{}} & 37.4 & 31.4 & 37.1 & 16.2 & 53.0 & 54.9 & 35.9 & \textbf{38.0} \\
\bottomrule
}
\par\medskip
\filltabular{7}{@{}l*{7}{c}@{}}{%
\toprule
\multicolumn{8}{c}{\textbf{(b) Perception and overall}} \\
\midrule
\textbf{Model} & AI2D & OCRBench & \shortstack{SEED-Bench-\\2-Plus} & MMBench & MMStar & \textbf{Avg.} & \textbf{Overall} \\
\midrule
Qwen2.5-VL-32B (\TVTFmt{}) & 78.27 & 83.50 & 70.66 & 80.50 & 64.13 & 75.41 & 63.96 \\
\rowcolor{colorhead}\hspace{0.8em}\textbf{\SLPFmt{}} & 84.4 & 85.8 & 71.8 & 86.1 & 67.7 & \textbf{79.2} & \textbf{66.1} \\
\midrule
MiniCPM-o 4.5 (\TVTFmt{}) & 80.28 & 81.00 & 64.73 & 84.52 & 65.40 & 75.19 & 62.47 \\
\rowcolor{colorhead}\hspace{0.8em}\textbf{\SLPFmt{}} & 81.4 & 82.3 & 65.7 & 85.5 & 67.7 & \textbf{76.5} & \textbf{63.6} \\
\midrule
Gemma-3-12B-IT (\TVTFmt{}) & 74.90 & 67.20 & 63.81 & 75.23 & 49.33 & 66.10 & 46.84 \\
\rowcolor{colorhead}\hspace{0.8em}\textbf{\SLPFmt{}} & 82.2 & 71.1 & 67.4 & 79.2 & 56.8 & \textbf{71.3} & \textbf{51.9} \\
\bottomrule
}
\end{table}

\section{Question-Guided Attention: Case-Study Protocol and All-Layer Analysis}
\label{app:attention_protocol}

The three examples in Figure~\ref{fig:dynamath_sample} compare \VTVFmt{} with \VVTFmt{} on \rev{Qwen2.5-VL-7B}, using the same prompt template and identical image copies.

We use prefill self-attention in the language decoder, with queries from the second image and keys from the first; attention inside the vision encoder is excluded. Each query row is normalized over the first image's keys, so the maps show how attention is distributed within the first image rather than its total mass. \rev{Following common practice in Transformer interpretability and multimodal information-flow analyses~\citep{jawahar2019bert,tenney2019bert,caron2021emerging,zhang2025crossmodal}, we} average over all \rev{28} heads, the final four layers (\rev{24--27}), and all second-image queries, and plot the per-patch difference between \VTVFmt{} and \VVTFmt{}. The outlined patches form the smallest set that accounts for 50\% of the total absolute difference.

\begin{figure}[!t]
\centering
\includegraphics[width=\linewidth]{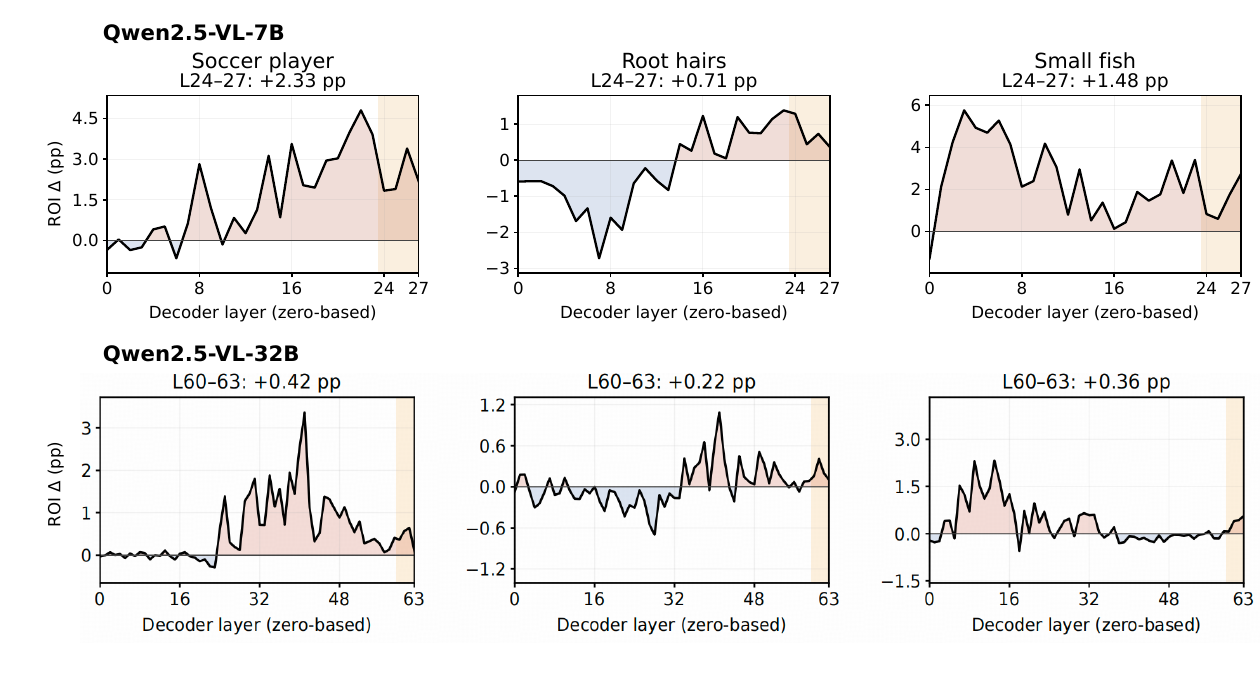}
\caption{\rev{Layer-wise answer-ROI attention share difference (\VTVFmt{}$-$\VVTFmt{}, percentage points) across all 28 decoder layers of Qwen2.5-VL-7B (top) and all 64 decoder layers of Qwen2.5-VL-32B (bottom) on the three Figure~\ref{fig:dynamath_sample} examples. Shaded orange bands mark the final four layers.}}
\label{fig:app_attention_curves}
\end{figure}

\begin{figure}[p]
\centering
\includegraphics[width=0.92\linewidth]{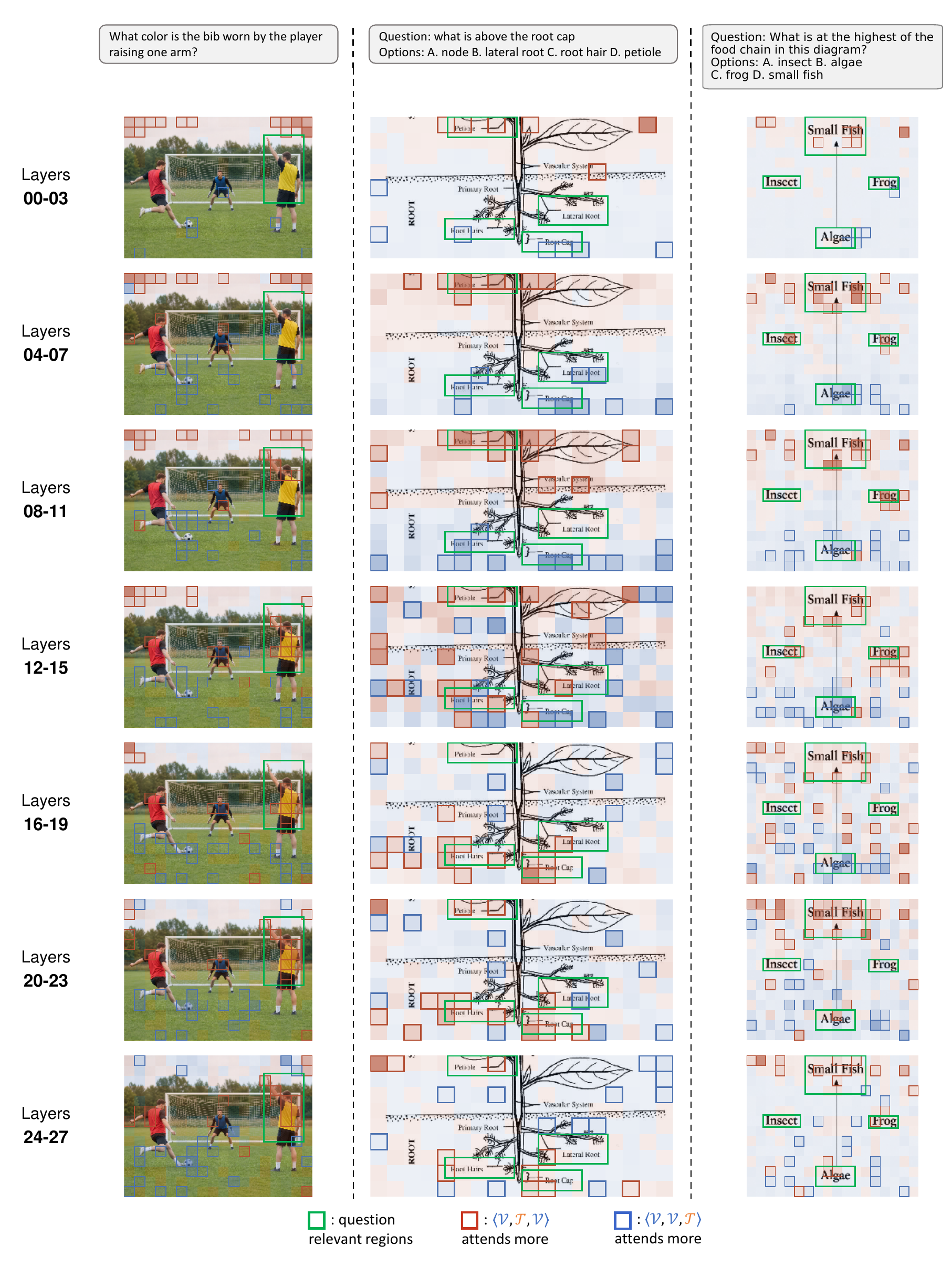}
\caption{\rev{Spatial attention-difference maps (\VTVFmt{}$-$\VVTFmt{}) across all seven consecutive four-layer windows of Qwen2.5-VL-7B on the three Figure~\ref{fig:dynamath_sample} examples.}}
\label{fig:app_attention_windows}
\end{figure}

\rev{\textbf{Sensitivity across decoder depth.}
To check whether the observation in Figure~\ref{fig:dynamath_sample} depends on selecting the final four layers, we compute answer-region attention share across all decoder layers, weighting boundary patches by their fractional overlap with a fixed answer-related region of interest (ROI). Table~\ref{tab:app_attention_windows} averages the \VTVFmt{}$-$\VVTFmt{} difference over consecutive four-layer windows on Qwen2.5-VL-7B, Figure~\ref{fig:app_attention_curves} plots per-layer ROI differences across all 28 layers of Qwen2.5-VL-7B and all 64 layers of Qwen2.5-VL-32B on the same three examples, and Figure~\ref{fig:app_attention_windows} shows the corresponding spatial maps across all seven windows of Qwen2.5-VL-7B. On Qwen2.5-VL-7B, every individual layer from 14 through 27 is positive in all three examples, and the four-layer average peaks in layers 20--23 rather than the final window; Qwen2.5-VL-32B exhibits a similar middle-to-late pattern.}

\begin{table}[!h]
\centering
\small
\setlength{\tabcolsep}{10pt}
\renewcommand{\arraystretch}{1.1}
\caption{\rev{Answer-region attention share difference (\VTVFmt{}$-$\VVTFmt{}, percentage points) across consecutive four-layer windows of Qwen2.5-VL-7B on the three Figure~\ref{fig:dynamath_sample} examples.}}
\label{tab:app_attention_windows}
\begin{tabular}{@{}lS[table-format=+1.2]S[table-format=+1.2]S[table-format=+1.2]@{}}
\toprule
\textbf{Layers} & {\textbf{Soccer}} & {\textbf{Root hairs}} & {\textbf{Food chain}} \\
\midrule
0--3   & -0.23 & -0.62 & +2.71 \\
4--7   & +0.23 & -1.68 & +4.76 \\
8--11  & +1.18 & -1.10 & +2.94 \\
12--15 & +1.35 & -0.17 & +1.41 \\
16--19 & +2.63 & +0.67 & +0.98 \\
20--23 & +3.94 & +1.01 & +2.60 \\
24--27 & +2.33 & +0.71 & +1.48 \\
\bottomrule
\end{tabular}
\end{table}

\FloatBarrier

\Needspace{10\baselineskip}
\section{Scratchpad Probe: Protocol and Additional Blocks}
\label{app:scratchpad}

\subsection{Protocol}

We use the strict single-answer multiple-choice subsets of DynaMath, We-Math, MMBench, MMStar, and AI2D, sampling 128 questions per benchmark for each model (640 per model). The probe prefills \(\langle\Vis_1,\Txt_1,\Rblk,\text{``Answer:''}\rangle\). Besides the image and question copies, \(\Rblk\) is a white or yellow image, \(1/f^2\) grayscale noise, repeated periods (Dots) or spaces, a fixed Lorem ipsum sequence, or a shuffled version of the same Lorem tokens. Within a model, all blocks except the question copy match the image copy in core token count and readout position. Noise and shuffled Lorem are averaged over three fixed seeds.

During readout, the forced ``Answer:'' cue attends only to \(\Rblk\) and itself, so it cannot bypass \(\Rblk\) to reach \(\Vis_1\) or \(\Txt_1\). We predict the option letter with the highest next-token log-probability. The probe gain is the accuracy gain over a cue-only baseline in which the cue is the only available context. This empirical baseline reflects the model's option-label preferences rather than uniform guessing.

\textbf{Blocked-context control.}
To test whether the gains depend on the earlier context, we additionally block \(\Rblk\)'s attention to all preceding keys during prefill. For every block in Table~\ref{tab:h3_carriers}, this returns accuracy to within 4.6 points of the cue-only baseline; the largest residual is for the yellow image on MiniCPM-o 4.5. Attention masks were verified with an independent implementation, and examples with different masks were never batched together.

\subsection{Shuffled Lorem Control}

Matching the token count does not make blocks interchangeable: on Qwen2.5-VL-32B, spaces yield $+40.9$ points, whereas Lorem yields $-0.9$ (Table~\ref{tab:h3_carriers}). Token order also matters. On Gemma-3-12B-IT, Lorem yields $+30.8$ points, but shuffling the same tokens reduces the gain to $-1.0$ (Table~\ref{tab:app_scratchpad_carriers}). Qwen2.5-VL-7B shows a smaller reduction, and on the other models Lorem is already ineffective.

\begin{table}[htbp]
\centering
\small
\setlength{\tabcolsep}{10pt}
\renewcommand{\arraystretch}{1.1}
\caption{Probe gains (points) for Lorem and a shuffled version with the same token multiset. Lorem values are repeated from Table~\ref{tab:h3_carriers}.}
\label{tab:app_scratchpad_carriers}
\begin{tabular}{@{}lS[table-format=+2.1]S[table-format=+1.1]@{}}
\toprule
\textbf{Model} & {\textbf{Lorem}} & {\textbf{Shuffled Lorem}} \\
\midrule
Qwen2.5-VL-3B  & -0.2 & -0.1 \\
Qwen2.5-VL-7B  & +8.3 & +3.1 \\
Qwen2.5-VL-32B & -0.9 & -1.3 \\
\midrule
MiniCPM-V 4.5 & -1.3 & -2.0 \\
MiniCPM-o 4.5 & +0.5 & -1.4 \\
\midrule
Gemma-3-12B-IT & +30.8 & -1.0 \\
\bottomrule
\end{tabular}
\end{table}

\subsection{Uniform Images Across Architectures}

In the scratchpad probe, white and yellow images yield at most $+2.2$ points on the three Qwen2.5-VL models but $+31.7$ to $+35.8$ points on MiniCPM-V/o 4.5 and Gemma-3-12B-IT (Table~\ref{tab:h3_carriers}). One possible explanation lies in where positional information enters the vision encoder. Gemma-3-12B-IT and MiniCPM-V/o 4.5 use SigLIP-based encoders that add learned absolute position embeddings to the patch embeddings before the first Transformer layer, so even a uniform image yields visual tokens that differ across positions.\footnote{See the \href{https://github.com/huggingface/transformers/blob/v5.17.0/src/transformers/models/siglip/modeling_siglip.py}{SigLIP embeddings}, and the MiniCPM-V 4.5 \href{https://huggingface.co/openbmb/MiniCPM-V-4_5/blob/11bfb79ddd18a4fed602f70dacd47a762ea63ab8/resampler.py}{resampler} and \href{https://huggingface.co/openbmb/MiniCPM-V-4_5/blob/d7bbb092045aef56fd040fdb1b9dd25d51337890/config.json}{configuration}.} MiniCPM's 64 learned resampler queries alone would not produce this variation: the resampler adds position only to its keys, so identical encoder features would yield identical outputs for every query. Qwen2.5-VL's vision encoder has no additive position embedding and injects position only through rotary embeddings on queries and keys.\footnote{See the \href{https://github.com/huggingface/transformers/blob/v5.17.0/src/transformers/models/qwen2_5_vl/modeling_qwen2_5_vl.py}{Qwen2.5-VL vision encoder}.} Rotary embeddings change attention weights but not the values being averaged, so identical patches remain identical through the encoder and the patch merger, and a uniform image enters the decoder as repeated copies of a single vector. Noise images, whose visual tokens are distinct, are effective on Qwen2.5-VL (Table~\ref{tab:h3_carriers}), consistent with this account. The analysis concerns image blocks only; text blocks follow a different embedding and position path and may behave differently. Testing whether identical visual tokens cause the near-zero gains requires measuring and intervening on visual-token states.

\subsection{\rev{Standard-Generation Carrier, Order, and Attention-Blocking Controls}}
\label{app:scratchpad_generation}

\rev{To compare restricted-probe readability with downstream utilization during standard generation, we evaluate six representative benchmarks (MATH-Vision, LogicVista, and DynaMath for reasoning; AI2D, OCRBench, and MMBench for perception) under matched decoding. Table~\ref{tab:app_gen_carriers} compares inserting the image copy (\SLPFmt{}), a white image (Blank), or length-matched spaces (\(\langle\Vis,\Txt,\text{Spaces},\Txt\rangle\)) relative to \VTTFmt{}. Even when Spaces achieves high probe readability, its normal-generation reasoning gains remain much smaller than those of readable visual blocks.}

\rev{To test whether question-conditioned encoding of \(\Vis_2\) contributes to downstream accuracy, we evaluate two complementary controls. First, on LogicVista with Qwen2.5-VL-32B ($n=447$), masking only \(\Vis_2\) from attending to \((\Vis_1,\Txt_1)\) with FlexAttention~\citep{dong2025flexattention} during standard generation (\(\Vis_2\)-no-read) lowers accuracy from $59.28\%$ under \SLPFmt{} to $55.93\%$ (vs.\ $56.82\%$ under \VTTFmt{}; Table~\ref{tab:app_noread}). All three arms are run in the same FlexAttention-based implementation, separate from the runs in Table~\ref{tab:app_gen_perbench}, so their LogicVista accuracies differ slightly. Because test-time attention masking can also introduce distribution shift, Table~\ref{tab:app_vvtt} evaluates an unmasked token- and prefill-compute-matched order control, \VVTTFmt{}, where native causal attention prevents \(\Vis_2\) from attending to \(\Txt_1\); \SLPFmt{} outperforms \VVTTFmt{} across all three models. Table~\ref{tab:app_gen_perbench} lists the per-benchmark accuracies behind Tables~\ref{tab:app_gen_carriers} and~\ref{tab:app_vvtt}.}

\begin{table}[htbp]
\centering
\small
\setlength{\tabcolsep}{7pt}
\renewcommand{\arraystretch}{1.1}
\caption{\rev{Scratchpad probe gains (Table~\ref{tab:h3_carriers}) alongside standard-generation accuracy changes over \VTTFmt{} (points) across three reasoning and three perception benchmarks.}}
\label{tab:app_gen_carriers}
\begin{tabular}{@{}llS[table-format=+2.1]S[table-format=+1.2]S[table-format=+1.2]@{}}
\toprule
\textbf{Model} & \textbf{Added block \(\Rblk\)} & {\textbf{Probe gain}} & {\textbf{Gen.\ \(\Delta\) (Reas.)}} & {\textbf{Gen.\ \(\Delta\) (Perc.)}} \\
\midrule
MiniCPM-o 4.5  & Copy \(\Vis\) (\SLP{}) & +34.5 & +0.94 & +0.83 \\
               & Blank                  & +35.8 & +1.08 & +0.08 \\
               & Spaces                 & +0.2  & +0.02 & +0.04 \\
\midrule
Gemma-3-12B-IT & Copy \(\Vis\) (\SLP{}) & +34.8 & +1.57 & +0.26 \\
               & Blank                  & +34.8 & +0.91 & -0.19 \\
               & Spaces                 & +29.2 & +0.42 & -0.15 \\
\midrule
Qwen2.5-VL-32B & Copy \(\Vis\) (\SLP{}) & +38.1 & +1.03 & +0.96 \\
               & Blank                  & -0.9  & -0.23 & +0.24 \\
               & Spaces                 & +40.9 & +0.16 & -0.50 \\
\bottomrule
\end{tabular}
\end{table}

\begin{table}[htbp]
\centering
\footnotesize
\begin{minipage}[t]{0.38\linewidth}
\vspace{0pt}
\centering
\caption{\rev{Targeted \(\Vis_2\) attention-blocking intervention on LogicVista (Qwen2.5-VL-32B, \%).}}
\label{tab:app_noread}
\setlength{\tabcolsep}{3.5pt}
\renewcommand{\arraystretch}{1.12}
\begin{tabular}{@{}ccc@{}}
\toprule
\textbf{\VTTFmt{}} & \textbf{\SLPFmt{}} & \textbf{\(\Vis_2\)-no-read} \\
\midrule
56.82 & 59.28 & 55.93 \\
\bottomrule
\end{tabular}
\end{minipage}\hfill
\begin{minipage}[t]{0.59\linewidth}
\vspace{0pt}
\centering
\caption{\rev{Unmasked equal-input-budget order control across six benchmarks (macro-average accuracy, \%).}}
\label{tab:app_vvtt}
\setlength{\tabcolsep}{2pt}
\renewcommand{\arraystretch}{1.12}
\begin{tabular*}{\linewidth}{@{\extracolsep{\fill}}lcccc@{}}
\toprule
\textbf{Model} & \textbf{\VTTFmt{}} & \textbf{\VVTTFmt{}} & \textbf{\SLPFmt{}} & \(\boldsymbol{\Delta}\) \\
\midrule
MiniCPM-o 4.5  & 67.90 & 68.48 & 68.79 & +0.31 \\
Gemma-3-12B-IT & 55.73 & 56.18 & 56.64 & +0.46 \\
Qwen2.5-VL-32B & 68.76 & 68.60 & 69.75 & +1.15 \\
\bottomrule
\end{tabular*}
\end{minipage}
\end{table}

\begin{table}[htbp]
\centering
\footnotesize
\renewcommand{\arraystretch}{1.08}
\caption{\rev{Per-benchmark standard-generation accuracy (\%) behind Tables~\ref{tab:app_gen_carriers} and~\ref{tab:app_vvtt}.}}
\label{tab:app_gen_perbench}
\filltabular{7}{@{}l*{7}{c}@{}}{%
\toprule
\textbf{Input} & \shortstack{MATH-\\Vision} & \shortstack{Logic\\Vista} & \shortstack{Dyna\\Math} & AI2D & OCRBench & MMBench & \textbf{Avg.} \\
\midrule
\multicolumn{8}{@{}l}{\textit{MiniCPM-o 4.5}} \\
\hspace{0.8em}\VTTFmt{} & 46.35 & 51.45 & 61.44 & 81.31 & 81.50 & 85.37 & 67.90 \\
\hspace{0.8em}\(\langle\Vis,\Txt,\text{Spaces},\Txt\rangle\) & 45.69 & 51.90 & 61.72 & 81.19 & 82.20 & 84.91 & 67.93 \\
\hspace{0.8em}\(\langle\Vis,\Txt,\text{Blank},\Txt\rangle\) & 45.53 & 55.03 & 61.94 & 81.87 & 81.50 & 85.06 & 68.49 \\
\hspace{0.8em}\VVTTFmt{} & 45.03 & 53.24 & 62.75 & 81.28 & 82.80 & 85.76 & 68.48 \\
\rowcolor{colorhead}\hspace{0.8em}\textbf{\SLPFmt{}} & 47.14 & 52.35 & 62.57 & 81.74 & 82.10 & 86.84 & 68.79 \\
\midrule
\multicolumn{8}{@{}l}{\textit{Gemma-3-12B-IT}} \\
\hspace{0.8em}\VTTFmt{} & 31.74 & 35.79 & 34.95 & 82.45 & 72.50 & 76.93 & 55.73 \\
\hspace{0.8em}\(\langle\Vis,\Txt,\text{Spaces},\Txt\rangle\) & 32.17 & 36.02 & 35.55 & 82.74 & 71.60 & 77.09 & 55.86 \\
\hspace{0.8em}\(\langle\Vis,\Txt,\text{Blank},\Txt\rangle\) & 30.66 & 36.47 & 38.08 & 82.45 & 70.60 & 78.25 & 56.08 \\
\hspace{0.8em}\VVTTFmt{} & 31.64 & 37.14 & 36.37 & 82.67 & 72.10 & 77.17 & 56.18 \\
\rowcolor{colorhead}\hspace{0.8em}\textbf{\SLPFmt{}} & 31.64 & 38.03 & 37.52 & 82.84 & 70.80 & 79.02 & 56.64 \\
\midrule
\multicolumn{8}{@{}l}{\textit{Qwen2.5-VL-32B}} \\
\hspace{0.8em}\VTTFmt{} & 38.55 & 57.94 & 67.13 & 80.73 & 83.90 & 84.29 & 68.76 \\
\hspace{0.8em}\(\langle\Vis,\Txt,\text{Spaces},\Txt\rangle\) & 38.62 & 58.39 & 67.09 & 79.53 & 83.90 & 83.98 & 68.58 \\
\hspace{0.8em}\(\langle\Vis,\Txt,\text{Blank},\Txt\rangle\) & 39.14 & 57.27 & 66.51 & 80.38 & 84.90 & 84.37 & 68.76 \\
\hspace{0.8em}\VVTTFmt{} & 38.78 & 56.60 & 67.43 & 81.31 & 84.30 & 83.20 & 68.60 \\
\rowcolor{colorhead}\hspace{0.8em}\textbf{\SLPFmt{}} & 38.82 & 60.18 & 67.70 & 81.83 & 86.00 & 83.98 & 69.75 \\
\bottomrule
}
\end{table}

\Needspace{8\baselineskip}
\section{Shifted-Image Control}
\label{app:shifted_copy}

Section~\ref{sec:summarizer} hypothesizes that an exact image copy draws \(\Vis_2\)'s attention toward \(\Vis_1\) at the expense of attention to the question and computation within \(\Vis_2\). We test this hypothesis by shifting \(\Vis_2\), measuring its effect on generation and on prefill attention.

\textbf{Generation.}
We cyclically shift \(\Vis_2\) by 7, 14, or 28 pixels in each of the four directions, which preserves all pixels while breaking their spatial correspondence with \(\Vis_1\). Table~\ref{tab:app_shift} reports the change relative to intact \SLPFmt{} under default prompts. Reasoning scores never decrease at the reported precision, whereas perception changes are small.

\begin{table}[htbp]
\centering
\small
\setlength{\tabcolsep}{8pt}
\renewcommand{\arraystretch}{1.1}
\caption{Score change (points) from shifting \(\Vis_2\), relative to intact \SLPFmt{} and averaged over four shift directions.}
\label{tab:app_shift}
\begin{tabular}{@{}llS[table-format=+1.1]S[table-format=+1.1]S[table-format=+1.1]S[table-format=+1.1]@{}}
\toprule
\textbf{Model} & \textbf{Suite} & {\textbf{7 px}} & {\textbf{14 px}} & {\textbf{28 px}} & {\textbf{Avg.}} \\
\midrule
Qwen2.5-VL-32B & Reasoning  & +0.4 & 0.0  & +0.4 & +0.3 \\
               & Perception & +0.2 & -0.1 & -0.1 & 0.0 \\
\midrule
Qwen2.5-VL-3B  & Reasoning  & +0.4 & +0.4 & +0.6 & +0.5 \\
               & Perception & +0.4 & +0.2 & +0.1 & +0.2 \\
\midrule
MiniCPM-o 4.5  & Reasoning  & +1.4 & +1.3 & +1.1 & +1.3 \\
               & Perception & +0.6 & +0.4 & +0.2 & +0.4 \\
\bottomrule
\end{tabular}
\end{table}

\textbf{Prefill attention.}
We probe Qwen2.5-VL-32B on 100 LogicVista prompts in \VTVFmt{}, shifting only \(\Vis_2\) to the right by 7, 14, or 28 processed pixels (half a ViT patch, one ViT patch, or one LLM visual token); the baseline uses an exact copy. For each visual query in \(\Vis_2\), we sum its last-layer attention mass over keys in \(\Vis_1\), \(\Txt_1\), and \(\Vis_2\) (causally accessible keys only), then average over queries and prompts. Masses are not renormalized over the three groups; other prompt tokens receive the remainder.

\begin{table}[htbp]
\centering
\small
\setlength{\tabcolsep}{10pt}
\renewcommand{\arraystretch}{1.1}
\caption{Change in last-layer attention mass (percentage points) from \(\Vis_2\) after rightward shifts, relative to the unshifted \VTVFmt{} baseline (Qwen2.5-VL-32B, 100 LogicVista prompts).}
\label{tab:app_shift_attention}
\begin{tabular}{@{}lS[table-format=+1.2]S[table-format=+1.2]S[table-format=+1.2]@{}}
\toprule
\textbf{Shift} & {\(\Vis_2\!\rightarrow\!\Vis_1\)} & {\(\Vis_2\!\rightarrow\!\Txt_1\)} & {\(\Vis_2\!\rightarrow\!\Vis_2\)} \\
\midrule
7 px  & -0.28 & +0.12 & +0.13 \\
14 px & -0.30 & +0.13 & +0.13 \\
28 px & -0.41 & +0.19 & +0.16 \\
\bottomrule
\end{tabular}
\end{table}

All three shifts move attention from \(\Vis_1\) toward the question and \(\Vis_2\) itself, and larger shifts produce larger changes (Table~\ref{tab:app_shift_attention}). Together with the generation gains, this supports the hypothesis, although the probe does not establish how the attention changes produce the score gains.

\section{Repetition under Direct and Default Prompts: Per-Benchmark Results}
\label{app:text_replay}

Tables~\ref{tab:app_text_replay} and~\ref{tab:qwen32b-cot} give the per-benchmark results behind Table~\ref{tab:cot_replay_interaction}; averages match that table after rounding. Within each pair, only the input changes, while the prompt template and decoding configuration stay fixed.

\textbf{Text-only model.}
We evaluate Qwen2.5-32B on OpenBookQA~\citep{mihaylov2018openbookqa}, GSM8K~\citep{cobbe2021gsm8k}, MATH-500~\citep{hendrycks2021math,lightman2023letsverify}, the text-only math split of OlympiadBench~\citep{he2024olympiadbench}, and MMLU-Pro~\citep{wang2024mmlupro}, where \QQ{} repeats the question verbatim. Repetition helps on every benchmark under both prompt settings, but the average gain shrinks from $+3.09$ with direct-answer instructions to $+0.67$ with default prompts.

\textbf{VLM.}
We evaluate Qwen2.5-VL-32B on the 12 main benchmarks, comparing \BaseFmt{} with \SLPFmt{}. The overall gain is $+1.3$ with direct-answer instructions and $+1.4$ with default prompts.

\begin{table}[htbp]
\centering
\footnotesize
\setlength{\tabcolsep}{6pt}
\renewcommand{\arraystretch}{1.1}
\caption{Question repetition with Qwen2.5-32B: accuracy of \Q{} and \QQ{} under direct-answer and default prompts.}
\label{tab:app_text_replay}
\begin{tabular*}{\linewidth}{@{\extracolsep{\fill}}llcccccc@{}}
\toprule
\textbf{Prompt} & \textbf{Input} & OpenBookQA & GSM8K & MATH-500 & OlympiadBench & MMLU-Pro & \textbf{Avg.} \\
\midrule
\multirow{3}{*}{Direct}
& \Q{}  & 93.80 & 39.80 & 40.00 & 20.46 & 57.67 & 50.35 \\
& \QQ{} & 95.40 & 47.31 & 44.40 & 22.18 & 57.88 & 53.43 \\
& \(\Delta\) & +1.60 & +7.51 & +4.40 & +1.72 & +0.21 & \textbf{+3.09} \\
\midrule
\multirow{3}{*}{Default}
& \Q{}  & 95.80 & 95.60 & 80.20 & 48.36 & 68.78 & 77.75 \\
& \QQ{} & 96.60 & 95.68 & 81.60 & 49.05 & 69.14 & 78.41 \\
& \(\Delta\) & +0.80 & +0.08 & +1.40 & +0.69 & +0.36 & \textbf{+0.67} \\
\bottomrule
\end{tabular*}
\end{table}

\begin{table}[htbp]
\centering
\footnotesize
\renewcommand{\arraystretch}{1.08}
\caption{Image--question repetition with Qwen2.5-VL-32B. Shaded rows use \SLPFmt{} instead of \BaseFmt{} under the prompt setting above; deltas on the averages are relative to \BaseFmt{}.}
\label{tab:qwen32b-cot}
\filltabular{8}{@{}l*{8}{c}@{}}{%
\toprule
\multicolumn{9}{c}{\textbf{(a) Reasoning}} \\
\midrule
\textbf{Prompt} & \shortstack{Dyna\\Math} & \shortstack{MATH-\\Vision} & \shortstack{Logic\\Vista} & \shortstack{Visual\\Puzzles} & MMMU & MathVista & We-Math & \textbf{Avg.} \\
\midrule
Direct & 44.8 & 31.8 & 44.5 & 37.0 & 60.3 & 69.0 & 40.5 & 46.8 \\
\rowcolor{colorhead} \hspace{0.8em}\textbf{+SLP} & 49.0 & 33.0 & 47.0 & 38.8 & 61.7 & 70.3 & 43.0 & \score{\textbf{49.0}}{+2.1} \\
\midrule
Default & 67.0 & 37.3 & 54.4 & 38.2 & 65.8 & 72.7 & 46.9 & 54.6 \\
\rowcolor{colorhead} \hspace{0.8em}\textbf{+SLP} & 67.5 & 38.8 & 57.7 & 40.8 & 67.7 & 74.9 & 49.4 & \score{\textbf{56.7}}{+2.1} \\
\bottomrule
}
\par\medskip
\filltabular{7}{@{}l*{7}{c}@{}}{%
\toprule
\multicolumn{8}{c}{\textbf{(b) Perception and overall}} \\
\midrule
\textbf{Prompt} & AI2D & OCRBench & \shortstack{SEED-Bench-\\2-Plus} & MMBench & MMStar & \textbf{Avg.} & \textbf{Overall} \\
\midrule
Direct & 86.7 & 85.4 & 73.5 & 86.8 & 68.5 & 80.2 & 60.7 \\
\rowcolor{colorhead} \hspace{0.8em}\textbf{+SLP} & 87.3 & 85.3 & 73.3 & 87.3 & 68.9 & \score{\textbf{80.4}}{+0.2} & \score{\textbf{62.1}}{+1.3} \\
\midrule
Default & 83.6 & 85.5 & 72.6 & 85.5 & 66.8 & 78.8 & 64.7 \\
\rowcolor{colorhead} \hspace{0.8em}\textbf{+SLP} & 84.4 & 85.8 & 71.8 & 86.1 & 67.7 & \score{\textbf{79.2}}{+0.4} & \score{\textbf{66.1}}{+1.4} \\
\bottomrule
}
\end{table}

\section{Controlled Continued Training: Full Setting and Results}
\label{app:training_study}

\subsection{Curriculum and Data Construction}

LLaVA-OneVision trains in stages~\citep{li2024llavaonevision}: a single-image phase on 3.2M instructions produces the released SI checkpoint, and the subsequent OneVision phase uses a 1.6M mixture of single-image, multi-image, and video data whose roughly 800K single-image component is a balanced, high-quality subset of the single-image data. We initialize from the public LLaVA-OneVision-7B-SI checkpoint and train on this single-image subset, keeping only strict single-turn, single-image examples (729{,}599 in total).

The two runs use the same sample-ID groups and targets. The Standard run serializes each example as \BaseFmt{}; the SLP run serializes it as \SLPFmt{} with verbatim image and question copies. Both supervise the answer once. The runs use different offline packing caps to fit the longer SLP sequences but share sample assignments, optimizer steps, data seed, and optimization hyperparameters (Table~\ref{tab:app_training_setting}).

\begin{table}[htbp]
\centering
\small
\renewcommand{\arraystretch}{1.12}
\caption{Training and evaluation settings for the controlled continued-training study.}
\label{tab:app_training_setting}
\begin{tabularx}{\linewidth}{@{}>{\raggedright\arraybackslash}p{0.21\linewidth} >{\raggedright\arraybackslash}X@{}}
\toprule
\textbf{Component} & \textbf{Setting} \\
\midrule
Initialization & \href{https://huggingface.co/llava-hf/llava-onevision-qwen2-7b-si-hf}{\texttt{llava-hf/llava-onevision-qwen2-7b-si-hf}} \\
Training data & 729{,}599 unique strict single-turn, single-image examples; identical sample-ID groups in both runs \\
Presentations & Standard: \BaseFmt{}; SLP: \SLPFmt{} with exact image and question copies; one supervised answer in both runs \\
Schedule & 5{,}000 optimizer steps; checkpoints and evaluation every 500 steps; seed and data seed 42 \\
Optimization & fused AdamW; learning rate $10^{-5}$ for non-vision parameters and $2\!\times\!10^{-6}$ for the vision tower; cosine decay; 3\% warmup; weight decay 0; gradient norm clipped at 1; sample-token-mean loss normalization \\
Batch and packing & 16 GPUs; one offline-packed microbatch per GPU; no gradient accumulation; 16 packed sequences and on average 145.92 examples per optimizer step; packing cap 32{,}075 tokens for Standard and 62{,}725 for SLP \\
Compute and numerics & 2 nodes $\times$ 8 NVIDIA A100-SXM4-80GB; FSDP2 HSDP (shard size 8, replicate degree 2); BF16 compute, TF32 enabled, FP32 master optimizer state; padding removed for variable-length FlashAttention-2 \\
Runtime & 14.2 hours for Standard and 32.0 hours for SLP \\
Evaluation & vLLM 0.6.3 with the OneVision post-norm parity fix; BF16, eager mode, greedy decoding; maximum context 32{,}768 and maximum generation 2{,}048 tokens \\
Benchmarks & The 12 main benchmarks (MathVista on \texttt{testmini}); macro averages weight benchmarks equally \\
\bottomrule
\end{tabularx}
\end{table}

\subsection{Checkpoint Trajectory}

Table~\ref{tab:app_training_trajectory} lists the values plotted in Figure~\ref{fig:training_study_single_column}. Step 0 evaluates the shared initialization under each presentation; later steps evaluate each run under its training presentation. Relative to step 0, Standard gains 0.55 points and SLP gains 1.97. The SLP$-$Standard gap moves from $-0.03$ to $+0.94$ on the reasoning suite and from $-1.81$ to $+0.25$ on the perception suite.

\begin{table}[htbp]
\centering
\small
\setlength{\tabcolsep}{9pt}
\renewcommand{\arraystretch}{1.08}
\caption{Matched-run trajectory (12-benchmark macro averages). Wins and ties count benchmarks on which SLP scores higher than and equal to Standard, respectively.}
\label{tab:app_training_trajectory}
\begin{tabular}{@{}rccS[table-format=+1.2]c@{}}
\toprule
\textbf{Step} & \textbf{Standard} & \textbf{SLP} & {\textbf{SLP$-$Standard}} & \textbf{Wins / ties} \\
\midrule
0       & 48.46 & 47.69 & -0.77 & 5 / 1 \\
500     & 48.17 & 48.25 & +0.08 & 7 / 0 \\
1{,}000 & 48.26 & 47.96 & -0.30 & 4 / 0 \\
1{,}500 & 47.34 & 47.57 & +0.23 & 6 / 0 \\
2{,}000 & 48.83 & 48.52 & -0.31 & 4 / 0 \\
2{,}500 & 48.72 & 49.32 & +0.59 & 8 / 0 \\
3{,}000 & 48.82 & 49.31 & +0.49 & 8 / 1 \\
3{,}500 & 48.94 & 49.67 & +0.72 & 7 / 0 \\
4{,}000 & 49.22 & 49.41 & +0.19 & 6 / 0 \\
4{,}500 & 49.02 & 49.45 & +0.43 & 7 / 0 \\
5{,}000 & 49.01 & 49.66 & +0.65 & 8 / 2 \\
\bottomrule
\end{tabular}
\end{table}

\subsection{Cross-Presentation Evaluation}

The trajectory varies the training and inference presentations together. At step 5{,}000, we additionally evaluate each checkpoint under both presentations (Table~\ref{tab:app_training_cross_format}; summarized in Table~\ref{tab:training_cross_format}). Under \BaseFmt{} inference, the two checkpoints are tied (49.00 vs.\ 49.01); under \SLPFmt{} inference, the SLP-trained checkpoint leads by 1.25 points. On 10 of 12 benchmarks, SLP training helps more under SLP inference than under Standard inference.

\begin{table}[htbp]
\centering
\small
\setlength{\tabcolsep}{8pt}
\renewcommand{\arraystretch}{1.08}
\caption{Per-benchmark results at step 5{,}000 for each training (rows of Table~\ref{tab:training_cross_format}) and inference presentation. Bold marks the highest macro average.}
\label{tab:app_training_cross_format}
\begin{tabular}{@{}lcccc@{}}
\toprule
& \multicolumn{2}{c}{\textbf{Standard inference}} & \multicolumn{2}{c}{\textbf{SLP inference}} \\
\cmidrule(lr){2-3}\cmidrule(l){4-5}
\textbf{Benchmark} & Std.-trained & SLP-trained & Std.-trained & SLP-trained \\
\midrule
MATH-Vision       & 18.5 & 18.9 & 18.6 & 18.5 \\
We-Math           & 21.6 & 23.6 & 20.0 & 22.8 \\
OCRBench          & 69.7 & 69.1 & 71.1 & 70.7 \\
AI2D              & 82.4 & 81.8 & 79.8 & 82.2 \\
DynaMath          & 31.2 & 31.3 & 32.5 & 33.2 \\
LogicVista        & 33.6 & 33.3 & 34.2 & 34.1 \\
VisualPuzzles     & 26.4 & 29.0 & 29.0 & 28.3 \\
MMMU              & 45.0 & 43.9 & 43.9 & 46.1 \\
MathVista         & 58.3 & 57.8 & 56.3 & 58.3 \\
MMBench           & 78.6 & 78.1 & 77.6 & 78.3 \\
MMStar            & 57.3 & 56.5 & 54.7 & 58.1 \\
SEED-Bench-2-Plus & 65.5 & 64.8 & 63.2 & 65.6 \\
\midrule
\textbf{Macro avg.} & 49.01 & 49.00 & 48.41 & \textbf{49.66} \\
\bottomrule
\end{tabular}
\end{table}

\section{Inference Latency and Generation Length}
\label{app:cost}

\subsection{Per-Model Latency}
\label{app:efficiency}

Latency is measured with vLLM on 4$\times$A800 GPUs with tensor parallelism 4. For each benchmark, we sample 10\% of the examples uniformly at random, giving 2{,}610 examples per model, and evaluate \BaseFmt{} and \SLPFmt{} on the same subset with default prompts. Table~\ref{tab:appendix_efficiency} reports mean per-example end-to-end latency, expanding the family averages in Table~\ref{tab:efficiency}. Changes range from $-0.49$ to $+0.46$ seconds: five models become slower, three become faster, and one is unchanged at the reported precision. These measurements reflect default serving behavior; attributing the changes to prefill and decoding would require separate stage timings.

\begin{table}[htbp]
\centering
\small
\setlength{\tabcolsep}{10pt}
\renewcommand{\arraystretch}{1.08}
\caption{Mean end-to-end latency (seconds) with default prompts. \(\Delta\) is \SLPFmt{} minus \BaseFmt{}.}
\label{tab:appendix_efficiency}
\begin{tabular}{@{}lS[table-format=1.2]S[table-format=1.2]S[table-format=+1.2]@{}}
\toprule
\textbf{Model} & {\textbf{\BaseFmt{}}} & {\textbf{\SLPFmt{}}} & {\textbf{\(\Delta\)}} \\
\midrule
Qwen2.5-VL-72B & 5.01 & 5.47 & +0.46 \\
Qwen2.5-VL-32B & 5.73 & 5.87 & +0.14 \\
Qwen2.5-VL-7B  & 0.95 & 1.00 & +0.05 \\
Qwen2.5-VL-3B  & 0.78 & 0.78 & +0.00 \\
\midrule
MiniCPM-o 4.5 & 4.42 & 4.32 & -0.10 \\
MiniCPM-V 4.5 & 5.20 & 5.36 & +0.16 \\
\midrule
Gemma-3-27B-IT & 3.96 & 3.47 & -0.49 \\
Gemma-3-12B-IT & 1.87 & 1.93 & +0.06 \\
Gemma-3-4B-IT  & 1.13 & 1.08 & -0.05 \\
\bottomrule
\end{tabular}
\end{table}

\subsection{\rev{Low-Resolution Second Look}}
\label{app:lowres}

\rev{Because \(\Vis_1\) already provides full-resolution visual details while \(\Vis_2\) serves as a question-conditioned scratchpad, we test downsampling only \(\Vis_2\) to one half (\texttt{SLP-half}) or one quarter (\texttt{SLP-quarter}) of its visual-token budget. Accuracy is evaluated on the full splits (six representative benchmarks for Qwen2.5-VL-32B; five challenging benchmarks in default non-thinking mode for Qwen3.8-27B), while end-to-end latency and time-to-first-token (TTFT) are timed side-by-side on the same H100 node (TP=2, batch size 1, fixed 10\% subset per benchmark). As shown in Table~\ref{tab:app_lowres}, coarsening \(\Vis_2\) monotonically reduces visual tokens, TTFT, and E2E latency while slightly increasing accuracy over full-resolution \SLP{}. Table~\ref{tab:app_lowres_perbench} gives the per-benchmark accuracies.}

\begin{table}[htbp]
\centering
\footnotesize
\setlength{\tabcolsep}{2pt}
\renewcommand{\arraystretch}{1.1}
\caption{\rev{Accuracy and inference cost of downsampled \(\Vis_2\) replay.}}
\label{tab:app_lowres}
\begin{tabular*}{\linewidth}{@{\extracolsep{\fill}}llrrrrrr@{}}
\toprule
\textbf{Model} & \textbf{Input} & \textbf{Vis.\ tok.} & \textbf{Acc.\ (\%)} & \(\boldsymbol{\Delta}\) \textbf{vs.\ \BaseFmt{}} & \textbf{TTFT (ms)} & \textbf{E2E (s)} & \textbf{E2E vs.\ \SLP{}} \\
\midrule
Qwen2.5-VL-32B & \BaseFmt{}           & 1,416 & 68.89 & {---} & 174.4 & 5.04 & $-5.4\%$ \\
               & \SLPFmt{}            & 2,833 & 69.75 & +0.86 & 285.0 & 5.32 & {---} \\
               & \texttt{SLP-half}    & 2,124 & 69.91 & +1.02 & 251.6 & 5.29 & $-0.7\%$ \\
               & \texttt{SLP-quarter} & 1,778 & 69.95 & +1.06 & 206.1 & 5.19 & $-2.6\%$ \\
\midrule
Qwen3.8-27B    & \BaseFmt{}           & 784   & 59.87 & {---} & 155.4 & 40.43 & $-8.0\%$ \\
               & \SLPFmt{}            & 1,568 & 62.55 & +2.68 & 193.6 & 43.96 & {---} \\
               & \texttt{SLP-half}    & 1,183 & 62.78 & +2.91 & 180.6 & 41.75 & $-5.0\%$ \\
               & \texttt{SLP-quarter} & 999   & 62.98 & +3.11 & 166.3 & 40.86 & $-7.0\%$ \\
\bottomrule
\end{tabular*}
\end{table}

\begin{table}[htbp]
\centering
\footnotesize
\renewcommand{\arraystretch}{1.08}
\caption{\rev{Per-benchmark accuracy (\%) behind Table~\ref{tab:app_lowres}.}}
\label{tab:app_lowres_perbench}
\filltabular{7}{@{}l*{7}{c}@{}}{%
\toprule
\multicolumn{8}{c}{\textbf{(a) Qwen2.5-VL-32B}} \\
\midrule
\textbf{Input} & \shortstack{MATH-\\Vision} & \shortstack{Logic\\Vista} & \shortstack{Dyna\\Math} & AI2D & OCRBench & MMBench & \textbf{Avg.} \\
\midrule
\BaseFmt{} & 37.27 & 57.72 & 67.09 & 81.61 & 85.30 & 84.37 & 68.89 \\
\rowcolor{colorhead}\SLPFmt{} & 38.82 & 60.18 & 67.70 & 81.83 & 86.00 & 83.98 & 69.75 \\
\texttt{SLP-half} & 39.21 & 59.28 & 67.92 & 81.67 & 86.60 & 84.75 & 69.91 \\
\texttt{SLP-quarter} & 41.09 & 57.49 & 67.62 & 81.93 & 86.50 & 85.06 & 69.95 \\
\bottomrule
}
\par\medskip
\filltabular{6}{@{}l*{6}{c}@{}}{%
\toprule
\multicolumn{7}{c}{\textbf{(b) Qwen3.8-27B (\texttt{non-thinking})}} \\
\midrule
\textbf{Input} & \shortstack{MATH-\\Vision} & \shortstack{Baby\\Vision} & \shortstack{RealWorld\\QA} & ERQA & CharXiv & \textbf{Avg.} \\
\midrule
\BaseFmt{} & 71.09 & 23.71 & 78.95 & 54.70 & 70.90 & 59.87 \\
\rowcolor{colorhead}\SLPFmt{} & 73.55 & 23.97 & 83.66 & 57.49 & 74.10 & 62.55 \\
\texttt{SLP-half} & 73.06 & 26.55 & 83.92 & 58.19 & 72.20 & 62.78 \\
\texttt{SLP-quarter} & 72.07 & 27.32 & 83.27 & 59.23 & 73.00 & 62.98 \\
\bottomrule
}
\end{table}

\FloatBarrier
\subsection{Generation Length}
\label{app:length}

Table~\ref{tab:appendix_length} reports mean generation length in characters with default prompts, averaged over the 12 main benchmarks. SLP shortens the output on seven of nine models; the two increases (Qwen2.5-VL-72B, $+9$ characters; MiniCPM-V 4.5, $+83$) are below 2\% of the \BaseFmt{} lengths. SLP's gains therefore do not come from longer generations. MiniCPM-V/o 4.5 lengths carry extra variance because their reasoning-suite runs use sampling (Appendix~\ref{app:decoding}).

\begin{table}[htbp]
\centering
\small
\setlength{\tabcolsep}{6pt}
\renewcommand{\arraystretch}{1.08}
\caption{Mean generation length (characters) with default prompts, averaged over the 12 main benchmarks. \(\Delta\) is \SLPFmt{} minus \BaseFmt{}.}
\label{tab:appendix_length}
\begin{tabular}{@{}lS[table-format=4.0]S[table-format=4.0]S[table-format=4.0]S[table-format=4.0]S[table-format=4.0]S[table-format=+2.0]@{}}
\toprule
\textbf{Model}
& {\textbf{\BaseFmt{}}}
& {\textbf{\VVTFmt{}}}
& {\textbf{\VTTFmt{}}}
& {\textbf{\VTVFmt{}}}
& {\textbf{\SLPFmt{}}}
& {\textbf{\(\Delta\)}} \\
\midrule
Qwen2.5-VL-72B & 562  & 550  & 592  & 573  & 571  & +9 \\
Qwen2.5-VL-32B & 1176 & 972  & 1232 & 1101 & 1166 & -10 \\
Qwen2.5-VL-7B  & 642  & 568  & 617  & 590  & 570  & -72 \\
Qwen2.5-VL-3B  & 493  & 440  & 468  & 482  & 466  & -27 \\
\midrule
MiniCPM-o 4.5 & 5494 & 5432 & 5518 & 5501 & 5465 & -29 \\
MiniCPM-V 4.5 & 8013 & 8021 & 7698 & 8034 & 8096 & +83 \\
\midrule
Gemma-3-27B-IT & 379  & 378  & 375  & 331  & 353  & -26 \\
Gemma-3-12B-IT & 389  & 383  & 362  & 358  & 320  & -69 \\
Gemma-3-4B-IT  & 432  & 417  & 340  & 358  & 354  & -78 \\
\bottomrule
\end{tabular}
\end{table}

\section{Duplicate-Image Stress Tests}
\label{app:duplicate_image_stress}

SLP presents the image as two separate visual-token blocks, which could lead a model to count instances once per copy, confuse spatial relations across copies, or emit less reliable box coordinates. We test these risks on CountQA~\citep{tamarapalli2025countqa}, SpatialMQA~\citep{liu2025spatialmqa}, and RefCOCOg~\citep{mao2016generation}, reporting accuracy for counting and spatial relations and Acc@0.5 for grounding (Table~\ref{tab:duplicate_image_stress}).

\textbf{Counting and spatial relations.}
Counting changes are mixed: five models improve, four decline, and the mean change is $-0.12$ points, so repetition shows no consistent degradation. SLP improves SpatialMQA accuracy on eight of nine models, by $+0.54$ points on average.

\textbf{Grounding.}
RefCOCOg is the only task with a consistent decline. Five of the six Qwen2.5-VL and MiniCPM-V/o 4.5 models decline, with changes from $-2.70$ to $+1.00$ points and a mean of $-1.11$. Gemma-3-IT rarely emits valid boxes and is excluded from this average. We list this decline as a limitation in Section~\ref{sec:limitations}.

\begin{table}[htbp]
\centering
\footnotesize
\setlength{\tabcolsep}{5pt}
\renewcommand{\arraystretch}{1.1}
\newcommand{\na}[1]{{\textcolor{gray}{\tablenum[table-format=2.2]{#1}}}}
\newcommand{\nad}[1]{{\textcolor{gray}{\tablenum[table-format=+1.2]{#1}}}}
\caption{Duplicate-image stress tests. Std.\ is \BaseFmt{}, SLP is \SLPFmt{}, and \(\Delta\) is their difference in points. Gemma-3-IT rarely emits valid boxes, so its RefCOCOg results (gray) are excluded from that average.}
\label{tab:duplicate_image_stress}
\begin{tabular}{@{}l
S[table-format=2.2]S[table-format=2.2]S[table-format=+1.2]@{\hspace{14pt}}
S[table-format=2.2]S[table-format=2.2]S[table-format=+1.2]@{\hspace{14pt}}
S[table-format=2.2]S[table-format=2.2]S[table-format=+1.2]@{}}
\toprule
& \multicolumn{3}{c}{\textbf{CountQA}} & \multicolumn{3}{c}{\textbf{SpatialMQA}} & \multicolumn{3}{c}{\textbf{RefCOCOg}} \\
\cmidrule(lr){2-4}\cmidrule(lr){5-7}\cmidrule(l){8-10}
\textbf{Model}
& {Std.} & {SLP} & {$\Delta$}
& {Std.} & {SLP} & {$\Delta$}
& {Std.} & {SLP} & {$\Delta$} \\
\midrule
Qwen2.5-VL-72B & 31.48 & 32.00 & +0.52 & 43.87 & 45.72 & +1.86 & 83.02 & 84.02 & +1.00 \\
Qwen2.5-VL-32B & 24.54 & 23.69 & -0.85 & 43.68 & 44.61 & +0.93 & 87.34 & 86.80 & -0.53 \\
Qwen2.5-VL-7B  & 21.20 & 20.55 & -0.65 & 37.36 & 38.20 & +0.84 & 86.26 & 85.07 & -1.20 \\
Qwen2.5-VL-3B  & 15.31 & 14.20 & -1.11 & 34.67 & 35.97 & +1.30 & 83.21 & 81.50 & -1.71 \\
\midrule
MiniCPM-o 4.5 & 17.28 & 17.80 & +0.52 & 38.48 & 40.06 & +1.58 & 35.12 & 33.60 & -1.52 \\
MiniCPM-V 4.5 & 28.14 & 25.65 & -2.49 & 40.33 & 40.52 & +0.19 & 78.42 & 75.72 & -2.70 \\
\midrule
Gemma-3-27B-IT & 18.46 & 19.24 & +0.79 & 38.85 & 38.94 & +0.09 & \na{0.06} & \na{0.49} & \nad{+0.43} \\
Gemma-3-12B-IT & 17.28 & 18.59 & +1.31 & 37.45 & 38.66 & +1.21 & \na{0.00} & \na{0.02} & \nad{+0.02} \\
Gemma-3-4B-IT  & 14.86 & 15.71 & +0.85 & 31.23 & 28.07 & -3.16 & \na{0.00} & \na{0.00} & \nad{+0.00} \\
\midrule
\textbf{Avg.} & 20.95 & 20.83 & -0.12 & 38.44 & 38.97 & +0.54 & 75.56 & 74.45 & -1.11 \\
\bottomrule
\end{tabular}
\end{table}

\section{\rev{Evaluations on Recent Open-Source and Reasoning VLMs}}
\label{app:new_models}

\rev{To verify whether recent open-source model advances or explicit long-chain reasoning budgets remove the cross-modal partial-observation gap, we evaluate Qwen3.5-35B-A3B and Gemma-4-E4B-it on the six-benchmark suite (MATH-Vision, LogicVista, DynaMath, AI2D, OCRBench, and MMBench), and evaluate Qwen3.8-27B across all four native reasoning modes (\texttt{non-thinking}, \texttt{low}, \texttt{medium}, and \texttt{xhigh}) on five challenging benchmarks from its official evaluation suite (MATH-Vision, BabyVision, RealWorldQA, ERQA, and CharXiv). Tables~\ref{tab:app_new_six} and~\ref{tab:app_qwen38} report per-benchmark accuracy and mean generated output tokens (including thinking and final answer). Across every model and reasoning budget, \SLPFmt{} yields consistent gains (+1.55 on Qwen3.5-35B-A3B, +4.04 on Gemma-4-E4B-it, and +2.17 to +2.91 on Qwen3.8-27B) while generating slightly fewer output tokens at each explicit thinking effort.}

\begin{table}[!h]
\centering
\footnotesize
\renewcommand{\arraystretch}{1.1}
\caption{\rev{Per-benchmark results on Qwen3.5-35B-A3B and Gemma-4-E4B-it across six representative benchmarks.}}
\label{tab:app_new_six}
\filltabular{7}{@{}l*{7}{c}@{}}{%
\toprule
\textbf{Model} & \shortstack{MATH-\\Vision} & \shortstack{Logic\\Vista} & \shortstack{Dyna\\Math} & AI2D & OCRBench & MMBench & \textbf{Avg.} \\
\midrule
Qwen3.5-35B-A3B (\BaseFmt{}) & 69.05 & 71.59 & 79.50 & 84.23 & 89.70 & 86.07 & 80.02 \\
\rowcolor{colorhead}\hspace{0.8em}\textbf{+SLP} & 70.69 & 75.17 & 83.49 & 84.23 & 91.70 & 84.13 & \score{\textbf{81.57}}{+1.55} \\
\midrule
Gemma-4-E4B-it (\BaseFmt{}) & 43.09 & 38.03 & 33.91 & 75.00 & 74.10 & 70.74 & 55.81 \\
\rowcolor{colorhead}\hspace{0.8em}\textbf{+SLP} & 45.69 & 45.19 & 40.54 & 78.21 & 75.40 & 74.07 & \score{\textbf{59.85}}{+4.04} \\
\bottomrule
}
\end{table}

\begin{table}[!h]
\centering
\footnotesize
\renewcommand{\arraystretch}{1.1}
\caption{\rev{Per-benchmark accuracy (\%) and macro-average total output tokens on Qwen3.8-27B across four native reasoning modes.}}
\label{tab:app_qwen38}
\filltabular{7}{@{}l*{7}{c}@{}}{%
\toprule
\textbf{Mode} & \shortstack{MATH-\\Vision} & \shortstack{Baby\\Vision} & \shortstack{RealWorld\\QA} & ERQA & CharXiv & \textbf{Avg.\ Acc.} & \textbf{Out.\ tokens} \\
\midrule
\texttt{non-thinking} (\BaseFmt{}) & 71.09 & 23.71 & 78.95 & 54.70 & 70.90 & 59.87 & 3,146 \\
\rowcolor{colorhead}\hspace{0.8em}\textbf{+SLP} & 73.55 & 23.97 & 83.66 & 57.49 & 74.10 & \score{\textbf{62.55}}{+2.68} & 3,154 \\
\midrule
\texttt{low} (\BaseFmt{}) & 81.28 & 38.40 & 84.31 & 65.85 & 78.10 & 69.59 & 1,374 \\
\rowcolor{colorhead}\hspace{0.8em}\textbf{+SLP} & 83.19 & 46.65 & 84.44 & 65.85 & 80.70 & \score{\textbf{72.17}}{+2.58} & 1,322 \\
\midrule
\texttt{medium} (\BaseFmt{}) & 82.24 & 41.24 & 83.66 & 65.16 & 77.60 & 69.98 & 1,886 \\
\rowcolor{colorhead}\hspace{0.8em}\textbf{+SLP} & 82.60 & 46.91 & 84.71 & 69.34 & 77.20 & \score{\textbf{72.15}}{+2.17} & 1,720 \\
\midrule
\texttt{xhigh} (\BaseFmt{}) & 84.77 & 46.13 & 84.97 & 67.25 & 79.20 & 72.46 & 4,405 \\
\rowcolor{colorhead}\hspace{0.8em}\textbf{+SLP} & 85.66 & 52.84 & 86.41 & 68.99 & 83.00 & \score{\textbf{75.38}}{+2.91} & 4,285 \\
\bottomrule
}
\end{table}

\end{document}